\documentclass[10pt,journal,compsoc]{IEEEtran}

\ifCLASSOPTIONcompsoc
\usepackage[nocompress]{cite}
\else
\usepackage{cite}
\fi

\usepackage{CJKutf8} 
\usepackage{bbding}
\usepackage{times}
\usepackage{epsfig}
\usepackage{graphicx}
\graphicspath{{figures/}}
\usepackage{amsmath}
\usepackage{amssymb}
\usepackage[colorlinks,linkcolor=red]{hyperref}

\usepackage[utf8]{inputenc} 
\usepackage[T1]{fontenc}    
\usepackage{url}  
\usepackage{booktabs}       
\usepackage{amsfonts}       
\usepackage{nicefrac}       
\usepackage{microtype} 
\usepackage{algorithm}
\usepackage{algpseudocode}
\usepackage{amsfonts,amssymb}
\usepackage{array}
\usepackage{bm}
\usepackage{multirow}

\usepackage{times} 
\usepackage{helvet} 
\usepackage{courier} 
\usepackage{amsmath}
\usepackage{footnote}
\usepackage{multirow}
\usepackage{array}
\usepackage{algorithm}
\usepackage{algpseudocode}
\usepackage{amsfonts,amssymb}
\usepackage{booktabs}
\usepackage{xcolor}
\usepackage{verbatim}
\usepackage{url}

\usepackage{ragged2e}

\usepackage{colortbl}
\usepackage{arydshln}
\usepackage{multirow}
\usepackage{multicol}
\usepackage{dashrule}
\usepackage{fancyhdr}

\definecolor{mygray}{gray}{.9}

\begin{document}
\begin{CJK}{UTF8}{gbsn}
\title{PSVMA+: Exploring Multi-granularity Semantic-visual Adaption for Generalized Zero-shot Learning}	
\author{Man Liu, Huihui Bai, Feng Li, Chunjie Zhang, Yunchao Wei, Meng Wang,~\IEEEmembership{Fellow,~IEEE}, Tat-Seng Chua, Yao Zhao,~\IEEEmembership{Fellow,~IEEE} 
\IEEEcompsocitemizethanks{\IEEEcompsocthanksitem Man Liu, Huihui Bai, Chunjie Zhang, Yunchao Wei, and Yao Zhao are with the Institute of Information Science, Beijing Jiaotong University, Beijing, China and Beijing Key Laboratory of Advanced Information Science and Network Technology, Beijing, China.
\IEEEcompsocthanksitem Feng Li and Meng Wang are with the Hefei University of Technology, Hefei, China.
\IEEEcompsocthanksitem Tat-Seng Chua is with the National University of Singapore, Singapore.} \thanks{Corresponding author: Huihui Bai (hhbai@bjtu.edu.cn); Feng Li (fengli@hfut.edu.cn)}
}

\markboth{Journal of \LaTeX\ Class Files,~Vol.~14, No.~8, August~2015}%
{Shell \MakeLowercase{\textit{et al.}}: Bare Demo of IEEEtran.cls for Computer Society Journals}

\IEEEtitleabstractindextext{%
\begin{abstract}
\justifying 
Generalized zero-shot learning (GZSL) endeavors to identify the unseen categories using knowledge from the seen domain, necessitating the intrinsic interactions between the visual features and attribute semantic features. However, GZSL suffers from insufficient visual-semantic correspondences due to the attribute diversity and instance diversity. Attribute diversity refers to varying semantic granularity in attribute descriptions, ranging from low-level (specific, directly observable) to high-level (abstract, highly generic) characteristics. This diversity challenges the collection of adequate visual cues for attributes under a uni-granularity. Additionally, diverse visual instances corresponding to the same sharing attributes introduce semantic ambiguity, leading to vague visual patterns. To tackle these problems, we propose a multi-granularity progressive semantic-visual mutual adaption (PSVMA+) network, where sufficient visual elements across granularity levels can be gathered to remedy the granularity inconsistency. PSVMA+ explores semantic-visual interactions at different granularity levels, enabling awareness of multi-granularity in both visual and semantic elements. At each granularity level, the dual semantic-visual transformer module (DSVTM) recasts the sharing attributes into instance-centric attributes and aggregates the semantic-related visual regions, thereby learning unambiguous visual features to accommodate various instances. Given the diverse contributions of different granularities, PSVMA+ employs selective cross-granularity learning to leverage knowledge from reliable granularities and adaptively fuses multi-granularity features for comprehensive representations. Experimental results demonstrate that PSVMA+ consistently outperforms state-of-the-art methods.

\end{abstract}
}

\maketitle

\IEEEdisplaynontitleabstractindextext

\IEEEpeerreviewmaketitle

\section{Introduction and Motivation}\label{sec:introduction}

\begin{figure*}[htb]
\centering
\includegraphics[width=0.9\linewidth]{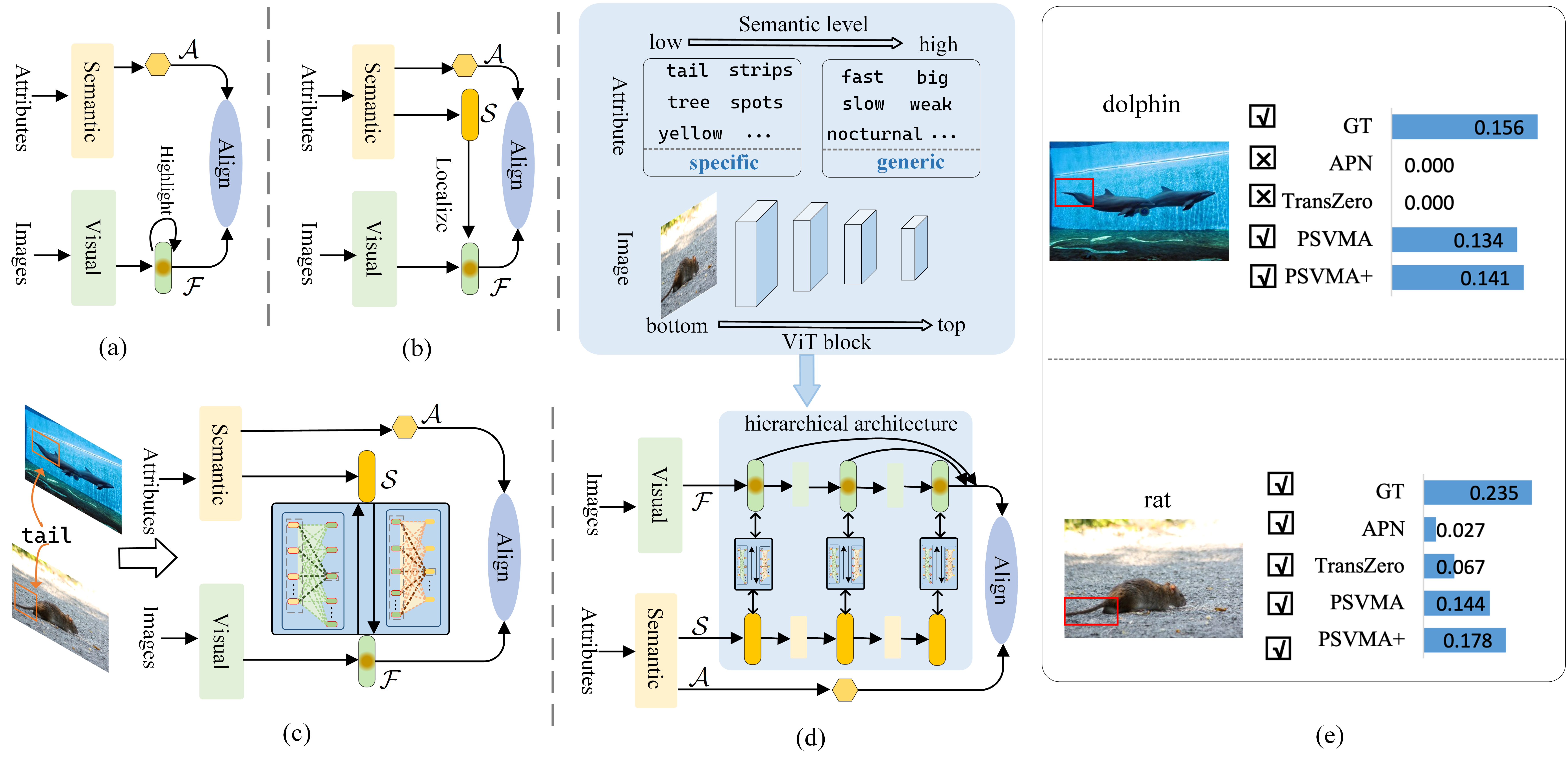}
\vspace{-5mm}
\caption{The embedding-based models and visualization for attribute disambiguation for GZSL. (a) Part-based methods via attention mechanisms. (b) Semantic-guided methods. (c) PSVMA. (d) PSVMA+. $\mathcal{A}$, $S$, $F$ denote the category attribute prototypes, sharing attributes, and visual features, respectively. The PSVMA+ achieves the best performances on (e) inter-class disambiguation.} 
\label{fig:motivation}
\end{figure*}

\IEEEPARstart{G}{eneralized} zero-shot learning (GZSL) \cite{palatucci2009zero} is a computer vision task that enables the recognition of both seen and unseen categories\footnote{It is noted that GZSL is typically denoted as generalized zero-shot image classification, and we follow the standard in this paper.}, using solely the seen data for training. Free from the requirement of enormous manually-labeled data, GZSL has extensively attracted increasing attention as a challenging recognition task that mimics human cognitive abilities \cite{Lampert2009LearningTD}. Early embedding-based approaches \cite{Xian2016LatentEF,zhang2017learning} involve the embedding of both the visual images and category attributes in a latent space to align global visual representations with their respective category prototypes. However, solely relying on this global information proves inadequate for fine-grained discriminative feature extraction, which is essential for capturing subtle differences between the seen and unseen classes. To address this issue, researchers have turned to part-based learning strategies for the exploration of distinctive local features to capture the subtle differences among the categories. Some prior works \cite{SGMA2019,AREN2019,RGEN2020} incorporate attention mechanisms into the networks to highlight the distinctive regions, as shown in Fig. \ref{fig:motivation} (a), where the modeling of the semantic-visual correspondence is only performed at the final category inference phase, that may cause biased recognition of seen classes. More recently, semantic-guided approaches~\cite{APN2020,MSDN2022} (\emph{cf} Fig. \ref{fig:motivation} (b)) make use of the sharing attributes to localize specific attribute-related regions. In this way, the correspondence is modeled during localization, which can effectively narrow the cross-domain gap to some extent.

Despite the promising performance of these methods, they still lack full consideration of the two key issues in GZSL:
\emph{i.e.,} the attribute diversity and instance diversity, which always play crucial roles in semantic-visual correspondence modeling. As a result, they suffer from ambiguous semantic predictions (\emph{cf} Fig. \ref{fig:motivation} (e)) owing to the insufficient alignment caused by the unmatched semantic-visual context. First, the diversity in attribute description poses a challenge in revealing sufficient visual clues under a uni-granularity space, which is commonly used in most previous methods~\cite{APN2020,MSDN2022,Chen2021TransZero}. As illustrated in Fig. \ref{fig:motivation} (d), attribute granularity refers to the semantic level of attribute descriptions, ranging from low- to high-level. Low-level attributes are specific and correspond to directly observable features like color, shape, or texture (e.g., ``tails,'' ``stripes,'' ``yellow''). These facilitate easier learning of visual-semantic associations. High-level attributes are more generic and abstract, often representing complex concepts or behaviors (e.g., ``fast,'' ``weak,'' ``nocturnal''). Therefore, neglecting these granularity disparities can lead to incorrect binding of the visual properties to semantic attributes. To comprehensively understand semantic knowledge, we argue that it is essential to exploit the granularity disparities of visual features and learn the semantic-visual interactions across granularities. Second, the diversity in visual appearances corresponding to the same sharing attribute descriptor may vary significantly across instances, leading to vague or even unrelated visual patterns for the given attribute. For example, as shown in Fig. \ref{fig:motivation} (c), for the attribute descriptor ``tail'', the visual depiction of a dolphin's tail is different from that of a rat's tail. Such an ambiguous relationship can cause deviation from the original visual semantics and result in misclassification in the unseen domain. Hence, it is desirable to recast the sharing attributes into visual-centric ones for more accurate correspondence.

To this end, we propose a multi-granularity progressive semantic-visual to mitigate the granularity disparities and semantic ambiguation, thereby improving the knowledge transferability for GZSL. As shown in Fig. \ref{fig:motivation} (d), PSVMA+ utilizes a hierarchical architecture to learn the granularity-wise semantic-visual alignment and cross-granularity visual clues aggregation. Specifically, considering the diversity of attributes, the visual object is first decomposed into multiple granularity layers, where different granularity visual features record the semantic representation of a certain attribute differently. At each granularity level, we devise a dual semantic-visual transformer module (DSVTM) comprising an instance-motivated semantic encoder and a semantic-motivated instance decoder. The encoder employs the instance-aware semantic attention to adapt the sharing attributes to various visual features. This allows DSVTM to convert the sharing attributes into instance-centric semantic features and recast the unmatched semantic-visual pair into the matched one. Subsequently, the decoder explores the correspondences between each visual patch and all matched attributes for semantic-related instance adaption, ultimately providing unambiguous visual representations of a specific granularity.

To make the best use of the multiple granularity-specific representations, directly collecting all of them is sub-optimal. This is because discriminative information varies in granularity, which contributes differently or could even be detrimental to category recognition. Hence, we present Selective Cross-Granularity Learning (SCGL) which consists of uncertainty-based granularity selection and cross-granularity learning. SCGL leverages highly reliable granularity representations as guidance to refine the worst ones while concentrating on challenging samples. Finally, we adopt Adaptive Multi-Granularity Fusion (AMGF) to leverage complementary multi-granularity features, distilling comprehensive representations for semantic knowledge transfer. In this way, PSVMA+ advances adequate semantic-visual interactions by leveraging the diversity of attributes and instances, facilitating more accurate inferences for both seen and unseen categories. Extensive experiments showcase that PSVMA+ achieves the new state-of-the-art on three GZSL benchmarks.

A preliminary version of this work was presented at a conference, referred to as PSVMA \cite{PSVMA2023} (\emph{cf} Fig. \ref{fig:motivation} (c)).
In this paper, we extend it in two major aspects.
i) We analyze the attribute and visual diversity in GZSL and strengthen the semantic-visual correlation in a multi-granularity manner by making full use of granularity disparities. 
ii) We apply SCGL to refine the representation of unreliable granularity and challenging samples. Moreover, the multi-granularity features are complementarily integrated to realize the acquisition of comprehensive semantic knowledge.

The main contributions of this paper are summarized as follows:
\begin{itemize}
\item We propose a PSVMA+ network to advance the GZSL from inconsistent semantic-visual information. It utilizes a hierarchical multi-granularity architecture for better semantic-visual alignment, accommodating various semantic attributes and visual instances.
\item We propose DSVTM to learn instance-centric attributes via an instance-motivated semantic encoder and model the accurate semantic-visual correspondences via a semantic-motivated instance decoder. The Hierarchical deployment captures granularity-specific unambiguous features.
\item We propose SCGL to enhance the representation of unreliable granularity and challenging samples under the guidance of reliable granularity. AMGF further enables collaborative gathering from diverse granularities, yielding comprehensive representations for GZSL.
\end{itemize}

The rest of this paper is organized as follows. Sec. \ref{sec2} discusses related works. The proposed PSVMA+ is illustrated in Sec. \ref{sec3}. Experimental results and discussions are provided in Sec. \ref{sec4}. Finally, we present a summary in Sec. \ref{sec5}.

\begin{figure*}[!htb]
\centering
\includegraphics[scale=0.58]{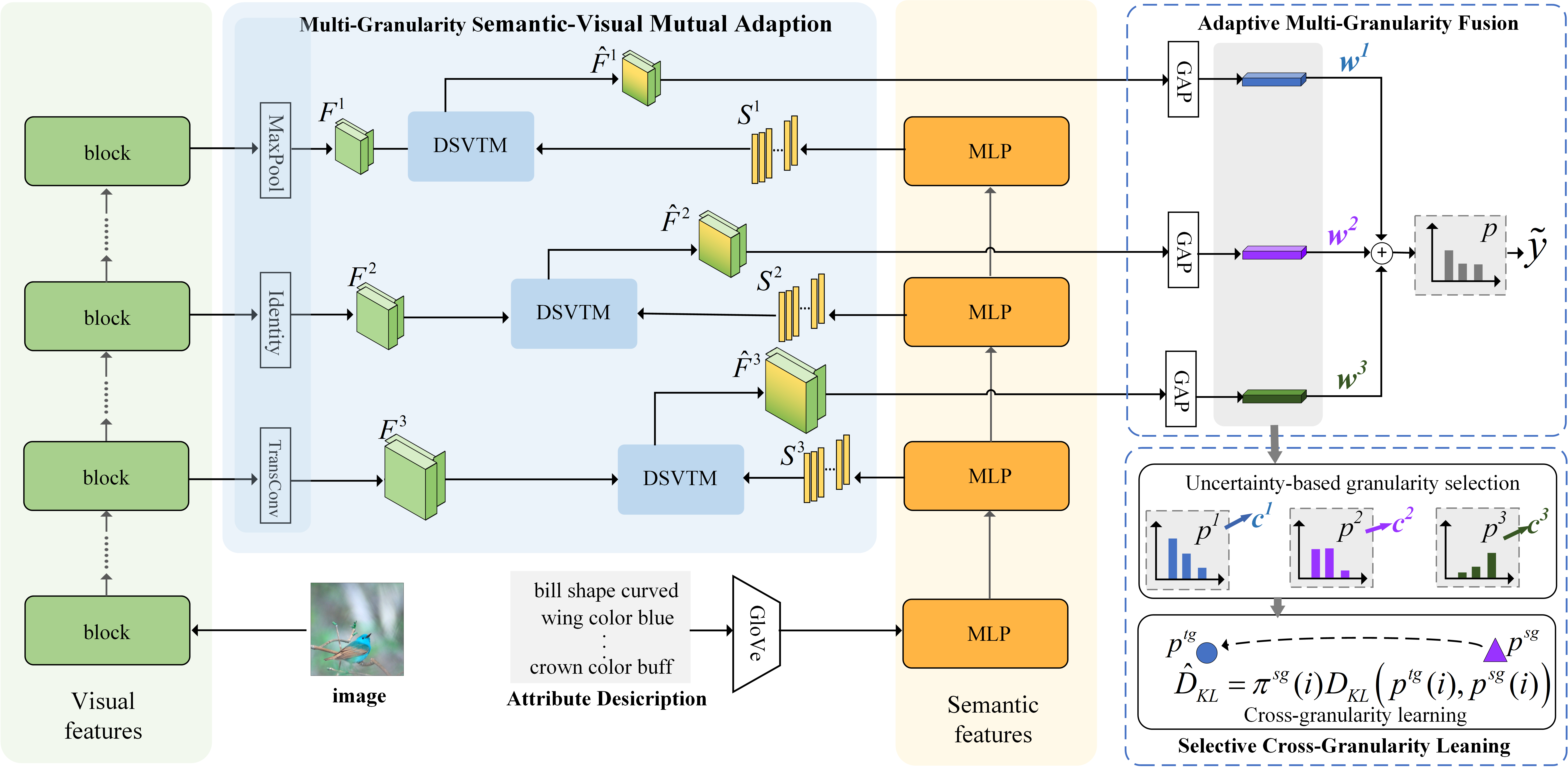}\vspace{-3mm}
\caption{The framework of our PSVMA+ model. PSVMA+ explores the semantic-visual interaction under a hierarchical multi-granularity architecture to model sufficient correspondence, mitigating the inconsistency caused by diverse attributes and instances. DSVTM conducts semantic-visual mutual adaptation under each granularity, yielding unambiguous granularity-specific visual features. Then, SCGL actively selects the reliable granularity to guide the refinement of the unreliable one, which also encourages the emphasis on the challenging samples positioned near the decision boundaries. Finally, AMGF augments the category decision-making process by dynamically fusing the multi-granularity features.}
\label{fig:framework}
\end{figure*}
\section{Related Work}\label{sec2}
In this section, we mainly review three streams of related works: generalized zero-shot learning, Transformer in GZSL, and uncertainty estimation.
\subsection{Generalized Zero-Shot Learning}\label{sec2.1} 
To transfer knowledge learned from the seen domain to the unseen domain, 
semantic information assumes a crucial role in providing a common space to describe seen and unseen categories. With the category attribute prototypes, generative GZSL approaches synthesize visual features of extra unseen categories by generative adversarial nets \cite{Huynh2020CompositionalZL,CEGZSL2021}, variational auto-encoders\cite{OCD2020,HSVA2021,SDGZSL2021}, or a combination of both \cite{TF-VAEGAN2020,FREE2021}. Although these methods compensate for the absence of the unseen domain during training, the introduction of extra data converts the GZSL problem into a fully supervised task.

The embedding-based method is the other mainstream branch for GZSL that projects and aligns the information originating from visual and semantic domains. Early works \cite{SJE2015,Xian2016LatentEF,zhang2017learning} directly map the global visual and semantic features into a common space for category predictions. Global visual information, however, falls short of capturing the subtle but substantial differences between the categories, weakening discriminative representations. To highlight discriminative visual regions, recent efforts have attempted the part-based techniques. Some works
\cite{LDF2018,SGMA2019} crop and zoom in on significant local areas employing coordinate positions obtained by attention mechanisms. Distinctive visual features are also emphasized by graph networks \cite{RGEN2020,hu2021graph} or attention guidance \cite{AREN2019,LFGAA2019,DVBE2020}. Furthermore, the sharing attribute prototypes, which are the same for all input images, have been introduced in semantic-guided methods \cite{APN2020,DPPN2021,MSDN2022} to localize attribute-related regions. Among these methods, DPPN\cite{DPPN2021} updates attribute prototypes to achieve superior performance. However, neglecting deeply mutual interactions between the semantic and visual domains, DPPN cannot alleviate the semantic ambiguity well.

\subsection{Transformers in GZSL}
Transformers \cite{Vaswani2017AttentionIA} have a strong track record of success in Natural Language Processing (NLP) and have gradually imposed remarkable achievements in computer vision tasks \cite{ViT2020,dong2022incremental}. Unlike CNNs, which are regarded as hierarchical ensembles of local features, transformers with cascaded architectures are encouraged to develop global-range relationships through the contribution of self-attention mechanisms. Despite the effectiveness of the transformer's architecture (such as the vision transformer (ViT)\cite{ViT2020}), research on GZSL has lagged behind, with just a small number of works \cite{ViT-ZSL2021,IEAM-ZSL2021,chen2022duet} using the ViT as a visual backbone. ViT-ZSL \cite{ViT-ZSL2021} directly aligns the patch tokens of ViT with the attribute information and maps the global features of the classification token to the semantic space for category prediction.
IEAM-ZSL \cite{IEAM-ZSL2021} not only captures the explicit attention by ViT, but also constructs another implicit attention to improve the recognition of unseen categories. DUET \cite{chen2022duet} proposes a cross-modal mask reconstruction module to transfer knowledge from the semantic domain to the visual domain. These works verify that, compared to CNNs, ViT specifically attends to image patches linked to category prototypes in GZSL. However, they neglect the semantic ambiguity problem and fail to construct matched semantic-visual correspondences, thus limiting their transferability and discriminability. 

\subsection{Uncertainty Estimation}
Uncertainty estimation in machine learning is crucial for quantifying the confidence of model predictions in complex decision-making scenarios. Early methods include Bayesian Neural Networks \cite{blundell2015weight}, which learn probability distributions over network weights. Evidential Deep Learning \cite{sensoy2018evidential} advances this by modeling uncertainty directly in class probabilities using subjective logic and Dirichlet distributions. Ensemble method \cite{lakshminarayanan2017simple} captures model uncertainty through aggregated predictions from multiple networks. Furthermore, the application of uncertainty estimation extends to active learning, where it guides sample selection strategies. Criteria such as entropy \cite{shannon2001mathematical}, confidence \cite{lewis1994heterogeneous}, and margin \cite{roth2006margin} have been employed to assess sample uncertainty, optimizing the learning process by focusing on the most informative instances. More recently, some works integrate the uncertainty estimation in zero-shot learning scenarios. \cite{chen2021entropy} attempts to minimize the uncertainty by entropy-based calibration in the overlapped areas of both seen and unseen classes. DCN \cite{Liu2018GeneralizedZL} also utilizes entropy as the measure of uncertainty, calibrating the confidence of source classes and the uncertainty of target classes simultaneously. DGCNet \cite{zhang2023dual} introduces dual uncertainty perception modules for visual and semantic features to explicitly quantify uncertainties. Different from these methods, in this paper, inspired by the uncertainty-based sample selection in active learning, we aim to choose the more reliable granularity for each sample based on their uncertainty, facilitating knowledge transfer from seen to unseen domains.

\section{Proposed Method}\label{sec3}
\noindent
\textbf{Problem Setting.} GZSL attempts to identify unseen categories by the knowledge transferred from the seen domain $\mathcal{D}^s$ to unseen domain $\mathcal{D}^u$. 
$\mathcal{D}^s=\{(x, y,a_y )|x \in \mathcal{X}^{s},y \in \mathcal{Y}^{s},a_y \in \mathcal{A}^{s}\}$, where $x$ refers to an image in $\mathcal{X}^{s} $, $y$ and $a_y$ refer to the corresponding label and category attributes. Here, $\mathcal{D}^u=\{(x^{u}, u,a_{u} )\}$, $x^{u} \in \mathcal{X}^{u}$, $u \in \mathcal{Y}^{u}$, $a_{u} \in \mathcal{A}^{u}$, and $\mathcal{A}=\mathcal{A}^{s} \cup \mathcal{A}^{u}$. The category space is disjoint between seen and unseen domains ($\mathcal{Y}^{s} \cap \mathcal{Y}^{u}=\varnothing$). In GZSL, the testing data includes both seen and unseen categories ($\mathcal{Y}=\mathcal{Y}^{s} \cup \mathcal{Y}^{u}$), while ZSL testing only involves unseen categories ($\mathcal{Y}=\mathcal{Y}^{u}$). Thus, an important problem is the seen-unseen bias, \emph{i.e.}, testing samples are more likely to be assigned to the seen categories observed during training. The goal of this work is to design an effective framework that explores semantic-visual interactions under the multi-granularity architecture.

\noindent
\textbf{Overview.}
In this work, we present the PSVMA+, a multi-granularity progressive semantic-visual mutual adaption network, where its overall pipeline is shown in Fig. \ref{fig:framework}. Given the image and sharing attribute description as input, PSVMA+ employs the vanilla ViT~\cite{ViT2020} and GloVe~\cite{Pennington2014GloveGV} followed by serial simplified MLP-mixer~\cite{tolstikhin2021mlp} to extract the visual ($F$) and semantic ($S$) features respectively. PSVMA+ takes the last $M$ blocks into consideration and establishes a hierarchical multi-granularity architecture, \emph{i.e.,} totally $M$ granularities, thereby leading to paired features $\{(F^1, S^1), (F^2, S^2), ..., (F^M, S^M)\}$, where each ViT block and MLP-mixer corresponds to a specific granularity (we use $M=3$ in this work for illustration). Notably, since ViT can only operate on the features at a single scale, to achieve multiple granularities, we append a multi-scale adaptor after these ViT blocks to achieve resolution adaption for later hierarchical DSVTMs. Compared with the original spatial size of ViT features, the multi-scale adaptor produces two other features of different resolutions $F^1$, $F^3$ by downsampling and upsampling with the scale factor 2.

Then, under each granularity, we devise the dual semantic-visual transformer module (DSVTM) that performs semantic-visual alignment and mutual adaption, yielding unambiguous attribute-related visual representations. With such granularity-specific representations, we propose selective cross-granularity learning (SCGL) that utilizes a uncertainty-based granularity selection strategy to steer the learning of unreliable granularity while focusing on the challenging samples positioned near the decision boundaries. Finally, an adaptive multi-granularity fusion (AMGF) is developed to better integrate the hierarchical information with different granularities in an element-wise convex weighted pattern.

\subsection{Dual Semantic-Visual Transformer}\label{sec3.1}
At each granularity, DSVTM takes $\{(F, S) \in \{(F^1, S^1), (F^2, S^2), (F^3, S^3)\}$ as inputs to achieve semantic-visual self-adaptation specific to that granularity. Assuming $F \in \mathbb{R}^{N_v \times D}$ and $S\in \mathbb{R}^{N_s \times D}$,where $N_v$ and $N_s$ denote the patch length and attribute prototypes with $D$ dimensional vectors, respectively.
\begin{figure*}[t]
\begin{center}
\includegraphics[scale=0.85]{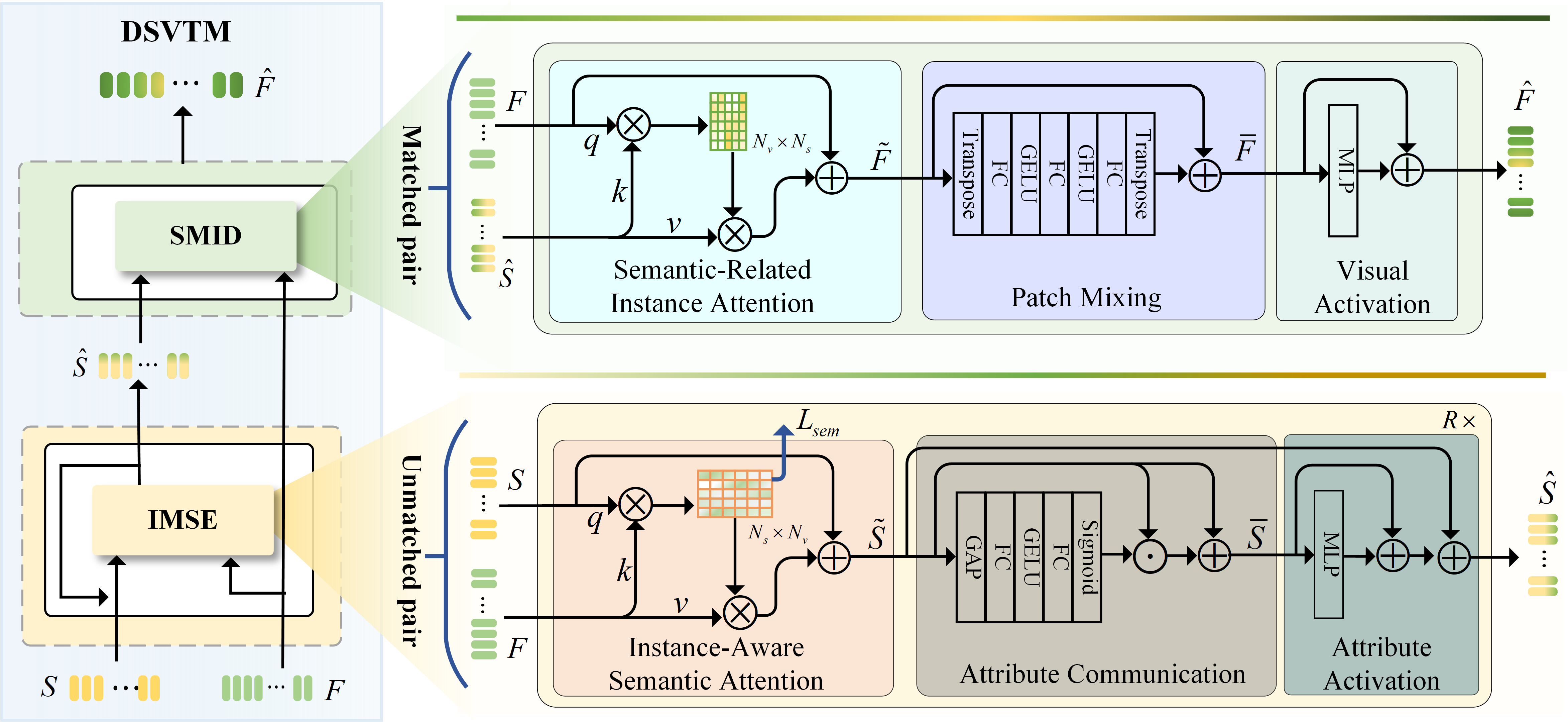}
\caption{DSVTM is a transformer-based structure that contains the IMSE and SMID, pursuing semantic and visual mutual adaption for the alleviation of semantic ambiguity. The IMSE in DSVTM progressively learns the instance-centric semantics to acquire a matched semantic-visual pair. The SMID in DSVTM constructs accurate interactions and learns unambiguous visual representations.}
\label{fig:psvma}
\end{center}
\end{figure*}

\subsubsection {Instance-Motivated Semantic Encoder}\label{sec3.1.1}
In DSVTM, the proposed instance-motivated semantic encoder (IMSE) aims at progressively learning instance-centric prototypes to produce the accurate matched semantic-visual pairs in a recurrent manner.
As shown in Fig. \ref{fig:psvma},
IMSE contains the instance-aware semantic attention, attribute communication, and activation, which are elaborated as follows.

\noindent
\textbf{Instance-Aware Semantic Attention.}
To adapt the sharing attributes $S$ to different instance features $F$, IMSE first executes cross-attention to learn attentive semantic representations based on instance features.
For the $i$-th attribute $S_i$, we search for the most relevant patch $F_j$ by modeling the relevance $\mathcal{M}_{(i,j)}$: 
\begin{equation}
\mathcal{M}_{(i,j)}=Q_{S_i}\cdot K_{F_j}^{\mathsf{T}}
\label{eq:M}
\end{equation}
\begin{equation}
\tilde S={\rm softmax}(\mathcal{M})\cdot V_{F} + S
\end{equation}
where $Q_{S_i}$, $K_{F_j}$, and $V_{F}$ are the query, key, and value learned from $S_i$ and $F_j$, $F$ respectively.
$\mathcal{M}_{(i,j)}$ indicates the localized region $F_{j}$ related to the attribute descriptor $S_{i}$, forming an affinity matrix
$\mathcal{M}\in \mathbb{R}^{N_s \times N_v}$. Then, $\mathcal{M}$ is applied to select distinct visual patches in $F$ related to each attribute,
packed together into instance-related attribute prototypes $\tilde{S} \in \mathbb{R}^{N_s \times D}$ with a residual connection.
To encourage the localization ability of patches related to attributes, 
we apply a semantic alignment loss to align $M$ with its category prototypes $a_{y}$:
\begin{equation}
\mathcal{L}_{sem}=\|{\rm GMP}(\mathcal{M})-a_y\|_{2}^{2}
\label{eq:L_sem}
\end{equation} 
where GMP is the 1-dimensional global max pooling operation. Thus, compared to the original sharing attribute prototype $S$, the instance-motivated semantic attribute $\tilde{S}$ is more closely linked to a specific instance.

\noindent
\textbf{Attribute Communication and Activation.}
As attribute descriptors are interdependent, IMSE then conducts attribute communication to compact the relevant attributes and scatter the irrelevant attributes via a group compact attention: 
\begin{equation}
\bar{S}= {\rm sigmoid}(\sigma ({\rm GMP}({\tilde{S}}) \cdot W_1)\cdot W_2)\cdot{\tilde{S}}+{S}
\label{eq:f_s}
\end{equation} 
where $\sigma$ is the GELU \cite{gelu} function. $W_1 \in \mathbb{R}^{N_s \times \frac{N_s}{\hbar}}$ and $W_2 \in \mathbb{R}^{\frac{N_s}{\hbar} \times N_s}$ are the parameters of two fully-connected (FC) layers, respectively. $\hbar$ denotes the number of attribute groups given in the datasets (\emph{e.g.}, 28 groups for 312 attributes on CUB dataset \cite{DatasetCUB}).
To make use of the compacted attribute prototypes $\bar{S}$, we further activate significant features and squeeze out trivial ones in each attribute by an MLP layer: 
\begin{equation}
\hat{S}={\rm MLP}(\bar{S})+ \bar{S} + {\tilde{S}}
\label{eq:mlp}
\end{equation} 
Here, cooperated with residual connections of $\bar{S}$ and $\tilde{S}$, more instance-aware information can be preserved.

Through repeated iterations of IMSE, we can progressively adapt the sharing attributes by observing previously adapted ones and visual instances,
mining instance-centric attribute prototypes.
With the produced $\hat{S}$, the unmatched semantic-visual pairs $(S, F)$ can be recast into matched pairs $(\hat{S}, F)$ ultimately.

\subsubsection{Semantic-Motivated Instance Decoder}\label{sec3.1.2}
Given the matched semantic-visual pair $(\hat{S}, F)$ from IMSE, 
we design a semantic-motivated instance decoder (SMID) to strengthen their interactions and learn unambiguous visual representations via the semantic-related instance attention, patch mixing and activation, as shown in Fig. \ref{fig:psvma}.

\noindent
\textbf{Semantic-Related Instance Attention.}
To acquire semantic-related visual representations, 
we first model the correspondence between the matched semantic-visual pair $(\hat{S}, F)$ via a cross-attention. 
Compared to the attention in IMSE (Eq. (\ref{eq:M})), here we focus on instance-centric attributes with respect to each visual patch and obtain attention weights $\bar{M}$, to select information in $\hat{S}$ and helps to aggregate significant semantic characteristics into the visual patches:
\begin{equation}
\tilde{F}={\rm softmax}(Q_F\cdot K_{\hat{S}}^{\mathsf{T}})\cdot V_{\hat{S}}+F
\label{eq:M'}
\end{equation} 
where $\tilde{F}$ denotes the visual instance representation that is aligned with the learned instance-centric attributes $\hat{S}$.
By such semantic-related attention, we can construct more accurate semantic-visual interactions and gather powerful matching attribute information in $\tilde{F}$. 

\noindent
\textbf{Patch Mixing and Activation.} 
Considering that the detailed information between different patches is essential for fine-grained recognition, 
we propose a patch mixing and activation module to expand and refine the association between patches.
Inspired by the concept of the manifold of interest in \cite{mobilenetv2}, we mix patches by an inverted residual layer with a linear bottleneck to improve the representation power.
This process can be formulated as:
\begin{equation}
F_{ex} = f_{ex}(({\tilde{F}})^{\mathsf{T}})=\sigma(({\tilde{F}})^{\mathsf{T}}\cdot W_{ex})
\label{eq:f_e}
\end{equation} 
\begin{equation}
F_{se} =f_{se}(F_{ex})=\sigma(F_ex\cdot W_{se})
\end{equation} 
\begin{equation}
F_{na} =f_{na}(F_{se})=F_{se}\cdot W_{na}
\end{equation} 
where $f_{ex}(\cdot)$ is an expansion layer consisting of an FC layer with parameters $W_{ex} \in \mathbb{R}^{N_v \times N_{h}}$ followed by an activation function.
Thus, the length of one patch is expanded to a higher dimension $N_{h}(N_{h}>N_v)$ for the subsequent information filtering implemented by a selection layer $f_{se}(\cdot)$.
Then, the selected patches are back-projected to the original low-dimension $N_v$ by a narrow linear bottleneck $f_{na}(\cdot)$. 
We utilize a shortcut to preserve complete information and produce $\bar{F} = (F_{na})^{\mathsf{T}} + \tilde F$. 
After that, the refined features in each visual patch are activated by an MLP layer with a residual connection:
\begin{equation}
\hat{F}={\rm MLP}(\bar{F})+ \bar{F}
\label{eq:hat_F}
\end{equation}

With SMID, we can take effects on the visual instance based on adapted semantic attributes along the spatial dimension, realizing the attentive visual features to keep with the attribute information. Moreover, equipped with our multi-granularity design, DSVTM enables to learn hierarchical unambiguous granularity-specific representations $\hat{F^{1}}$, $\hat{F^{2}}$, $\hat{F^{3}}$ for the inputs $\{(F^1, S^1), (F^2, S^2), (F^3, S^3)\}$.

\subsection{Selective Cross-Granularity Learning}\label{sec3.2} 
Based on learned granularity-specific representations $\hat{F^{1}}$, $\hat{F^{2}}$, $\hat{F^{3}}$, it is sub-optimal to directly combine all of them since the discriminative information varies in granularities, which contributes differently or could even be detrimental to category recognition.

Therefore, we design SCGL, comprising of uncertainty-based granularity selection and cross-granularity learning, to exploit the representations of high reliable granularity as guidance to refine the worst one. This also encourages the model to focus on the challenging samples positioned near the decision boundaries.

\subsubsection{Uncertainty-based Granularity Selection}\label{sec3.2.1}
To achieve granularity selection, one important thing is to determine the reliability of the features from different granularities, where the more reliable granularity contains more discriminative characteristics for category decisions. To this end, we first calculate the evaluation of category distribution. For the $g$-th granularity of the $i$-th visual input, we can obtain the distribution as:
\begin{equation}
\begin{gathered}
\mathcal{P}\left(\hat{y}_i=k \mid \hat{F}^{g}(i)\right)=\frac{\exp \left( \cos \left(\phi(\hat{F}^{g}(i)), a_{k}\right)/\tau\right)}{\sum_{\hat{k} \in \mathcal{Y}^S} \exp \left( \cos \left(\phi(\hat{F}^{g}(i)), a_{\hat{k}}\right)/\tau\right)}, \\
k \in \mathcal{Y}^S , \quad g \in\{1, 2,3\},
\end{gathered}
\label{eq:p}
\end{equation}
where $a_{k}$ and $\tau$ denote the $k$-th category prototypes and a temperature, respectively. $cos(\cdot)$ denotes the cosine similarity. $\phi(\cdot)$ is a mapping function that projects the granularity-specific visual features into the semantic space through average pooling and a linear layer, yielding predicted semantic attributes. Then, inspired by the uncertainty-based active learning algorithms~\cite{10203554},we apply absolute certainty to estimate the granularity reliability. Specifically, we define the absolute certainty $\mathcal{C}^{g}(i)$ as the maximum element within granularity-specific category distribution:
\begin{equation}
\mathcal{C}^{g}(i)=\max_{k}\mathcal{P}\left(\hat{y}_i=k \mid \hat{F}^{g}(i)\right)
\label{eq:certainty}
\end{equation}
where the higher absolute certainty represents the greater reliability of granularity and the ability to express discriminative recognition characteristics in GZSL. Conversely, when the certainty is low, the granularity is likely to be unreliable for recognition.
To facilitate the subsequent exploration of cross-granularity learning, we select the granularity with the largest reliability as $tg(i)=\arg \max {\mathcal{C}^{g}(i)}$ and the granularity with the lowest reliability is defined as $sg(i)=\arg \min {\mathcal{C}^{g}(i)}$.

\subsubsection{Cross-Granularity Learning}\label{sec3.2.2}
With the selected reliable granularity $tg(i)$, we propose a cross-granularity learning that adopts it as guidance to improve the representation learning for unreliable granularity $sg(i)$. An option is to conduct mutual knowledge distillation between each other, which selects arbitrary granularity no matter the reliable or unreliable one as the teacher and another as the student. However, our experiment shows that it is inferior as it will bring detrimental influence on reliable granularity when the unreliable one serves as the teacher (\emph{cf} Table~\ref{table:ablation_ge}). Besides, due to the varying influence of specific granularities on different samples, it is desirable to dynamically assign the most reliable granularity $tg(i)$ as the teacher and unreliable $sg(i)$ one as the student for each sample during knowledge distillation. 

Therefore, for the unreliable granularity $sg(i)$, we conduct granularity-aware Kullback-Leibler (KL) divergence between $tg(i)$ and $sg(i)$, formulated as:
\begin{equation}
\hat{D}_{K L}=\pi^{sg}(i) D_{K L}\left(p^{tg}(i), p^{sg}(i)\right)
\label{eq:kl}
\end{equation}
where $p^{tg}(i)$ (or $p^{sg}(i)$) denotes the granularities-specific distribution for the $tg$-th (or $sg$-th) granularity of the $i$-th sample as defined in Eq. (\ref{eq:p}). $\mathcal{D}_{K L}\left(p^{tg}(i), p^{sg}(i)\right)=p^{tg}(i) \log ({p^{tg}(i)}/{p^{sg}(i)})$. Utilizing the consistency constraint $D_{KL}$, the distillation strategy facilitates teacher-student ($p^{tg}(i)$-$p^{sg}(i)$) learning throughout the training phase. This process optimizes the unreliable granularity $sg(i)$ to learn from the reliable counterpart $tg(i)$ by mimicking the discriminative characteristics of $tg(i)$, thus enhancing its classification capability.
Furthermore, to encourage the network to prioritize challenging samples, we reweight the entire constraint using the coefficient $\pi^{sg}(i)$ which is defined as:
\begin{equation}
\pi_{sg}(i)=H\left(p^{sg}(i)\right)
\end{equation}
where $H$ is the information entropy to obtain $\pi^{sg}(i)$ as samples with higher entropy that are closer to the decision boundaries and are more prone to be misclassified.

\subsection{Adaptive Multi-Granularity Fusion}\label{sec3.2.3}
Since different categories possess distinct discriminative granularities, which often contribute differently to category recognition, we exploit the complementarity of multi-granularity features and distill comprehensive representations for semantic knowledge transfer. 
Here, each granularity is assigned a weight $w^{g}(i)$ that is determined by the absolute certainty in Eq. (\ref{eq:certainty}). Then, we sum up the granularity-specific features according to these weights to derive the comprehensive multi-granularity representation $F_A$. The process is formulated as:
\begin{equation}
w^{g}(i)=\frac{\mathcal{C}^{g}(i)}{\sum_{g=1}^3 \mathcal{C}^{g}(i)}
\end{equation}
\begin{equation}
F_A=\sum_{g=1}^3 w^{g}(i) \hat{F}^{g}(i)
\end{equation}
Subsequently, we map $F_A$ to the semantic space and calculate the semantic scores through the measurement of cosine similarity with category prototypes.
\begin{equation}
score(\hat{y}|x) =\cos \left(\phi(F_A), a_{\hat{y}}\right)
\label{eq:score}
\end{equation} 
The category of instance $x$ is supervised by the classification loss ${\cal L}_{cls}$ defined as:
\begin{equation}
{{\cal L}_{cls}} = - \log \frac{{\exp \left( {score(y|x)} \right)}}{{\sum\limits_{\hat y \in {{\cal Y}^S}} {\exp } \left( {score(\hat y|x)} \right)}}
\end{equation}

\subsection{Model Optimization and Inference}\label{sec3.4}
\noindent
\textbf{Optimization.}
In addition to the semantic alignment loss, granularity-aware KL loss, and the classification loss mentioned above, we design a debiasing loss $\mathcal{L}_{deb}$ to mitigate the seen-unseen bias.
To better balance the score dependency in the seen-unseen domain, $\mathcal{L}_{deb}$ is proposed to pursue the distribution consistency in terms of mean and variance:
\begin{equation}
{{\cal L}_{deb}}{\rm{ }} = \|{\alpha _s} - {\alpha _u}\|_2^2 + \|{\beta _s} - {\beta _u}\|_2^2
\label{eqlbias}
\end{equation}
$\alpha _s$ and $\beta _s$ denote the mean and variance value of seen predictions $score(\hat{y_s}|x, \hat{y_s} \in \mathcal{Y}^{s})$.

Finally, the overall optimization goal can be defined as:
\begin{equation}
\mathcal{L}=\mathcal{L}_{cls} +\lambda _{sem}\mathcal{L}_{sem}+\lambda_{KL}\hat{D}_{KL}+ \lambda_{deb}\mathcal{L}_{deb}
\label{eq:loss}
\end{equation}
where $\lambda _{sem}$, $\lambda _{KL}$ and $\lambda _{deb}$ are the hyperparameters that control the weights of semantic alignment loss $\mathcal{L}_{sem}$, granularity-aware KL loss $\hat{D}_{K L}$ and the debiasing loss $\mathcal{L}_{deb}$, respectively.

\noindent
\textbf{Inference.} 
During training, the model merely learns about the knowledge of seen categories, whereas both seen and unseen categories are contained at inference time under the GZSL setting.
Therefore, calibrated stacking (CS) \cite{2016cs} is applied to jointly define the category:
\begin{equation}
\tilde{y}=\arg \max _{\hat{y} \in  \mathcal{Y}}\left(score(\hat{y}|x)-\gamma\mathbb{I}_{\left[\hat{y} \in \mathcal{Y}^s\right]}\right)
\label{eq:pre}
\end{equation}
$\mathbb{I}_{\mathcal{Y}^{S}}(\cdot)$ denotes an indicator function, whose result is 1 when $\hat{y} \in \mathcal{Y^S}$ and 0 otherwise.
A calibrated factor $\gamma$ is applied to trade-off the calibration degree on seen categories and decides the category $\tilde{y}$ of a sample $x$. To provide a detailed understanding of our PSVMA+, the process, encompassing both model training and inferring, is outlined in the pseudocode presented in Algorithm \ref{algotithm:PSVMAp}.

\begin{algorithm}[t]
\caption{The algorithm of PSVMA+.}
\label{algotithm:PSVMAp}
\begin{algorithmic}[1]
\Require The dataset includes the seen data $\mathcal{D}^s=\{(x, y,a_y )|x \in \mathcal{X}^{s},y \in \mathcal{Y}^{s},a_y \in \mathcal{A}^{s}\}$, and the unseen data $\mathcal{D}^u=\{(x^{u}, u,a_{u} )\}$.
\Ensure The predicted label $\tilde{y}$ for the input samples under the maximum iteration $\mathrm{max}_{iter}$. 
\While{$iter \leq \mathrm{max}_{iter}$} 	\Comment{\textit{\color{gray} Optimization}}
\State Extract multi-granularity visual features $\{F^1, F^2, F^3\}$
\State Extract multi-granularity semantic features $\{S^1, S^2, S^3\}$
\While{$g \leq 3$}
\State {$\hat{F^{g}} \gets {DSVTM(F^g,S^g)}$}
\EndWhile
\State Select granularities ($sg,tg$) based on Eq. (\ref{eq:certainty}).
\State Refine granularities $sg$ by Eq. (\ref{eq:kl})
\State {$F_A \gets AMGF({\hat{F^1}},\hat{F^2},\hat{F^3})$}
\State {Mapping and classification.}
\State Optimize PSVMA+ by Eq. (\ref{eq:loss})
\EndWhile
\State Predict $\tilde{y}$ using Eq. (\ref{eq:pre}) \Comment{\textit{\color{gray} Prediction}}
\end{algorithmic}
\end{algorithm}

\section{Experiments}\label{sec4}
\noindent
\textbf{Datasets.}
We evaluate PSVMA+ on three benchmark datasets, \emph{i.e.}, Caltech-USCD Birds-200-2011 (CUB) \cite{DatasetCUB}, SUN Attribute (SUN) \cite{DatasetSUN}, Animals with Attributes2 (AwA2) \cite{DatasetAWA2}. The seen-unseen category division is set according to Proposed Split (PS) \cite{DatasetAWA2}. As shown in Table \ref{Tab:PS}, the CUB dataset comprises 11,788 images portraying 200 bird classes, with a distribution of 150/50 for seen/unseen classes, and it is characterized by 312 attributes. SUN, a large scene dataset, encompasses 14,340 images across 717 classes, divided into seen/unseen classes at 645/72, and annotated with 102 attributes. AwA2, while coarser with only 50 animal classes (seen/unseen classes = 40/10), is impressively rich in images (boasting 37,322 in total) and described by 85 attributes.

\noindent
\textbf{Metrics.}
We assess top-1 accuracy in both the ZSL and GZSL settings. In the ZSL setting, we only evaluate the accuracy on unseen classes, denoted as $acc$. In GZSL setting, we follow \cite{DatasetAWA2} and apply the harmonic mean (defined as $H=2 \times S \times U /(S+U)$)
to evaluate the performance of our framework. $S$ and $U$ denote the top-1 accuracy of the seen and unseen classes, respectively.

\noindent\textbf{Implementation Details.}
Unlike previous GZSL works that utilize ResNet\cite{Resnet2016} models as visual backbones, we harness the ViT-Base \cite{ViT2020} model \footnote{ImageNet-1k weights (pretrained on ImageNet-21k FT on ImageNet-1k): \url{https://github.com/WZMIAOMIAO/deep-learning-for-image-processing/blob/master/pytorch_classification/vision_transformer/vit_model.py}} as the visual feature extractor. The input image resolution is $224 \times 224$ and the patch size is $16 \times 16$. Our framework is implemented with Pytorch over an Nvidia GeForce RX 3090 GPU.

\begin{table}[ht]
\centering
\caption{Proposed Split (PS) of the benchmark datasets to evaluate our network. $N_s$ and $\hbar$ denote the number of attribute dimensions and attribute groups, respectively. $s$|$u$ is the number of seen|unseen classes.}
\resizebox{1.0\linewidth}{!}{
{
\begin{tabular}{c|c|c|c}
 \hline
 Datasets &classes ($s$ $|$ $u$) &images &$N_s$ ($\hbar$) \\ \hline
 CUB \cite{DatasetCUB} &200 (150 $|$ 50) &11,788 &312 (28) \\
 SUN \cite{DatasetSUN} &717 (645 $|$ 72) &14,340 &102 (4) \\
 AwA2 \cite{DatasetAWA2} &50 (40 $|$ 10) &37,322 &85 (9) \\ \hline
\end{tabular}}}
\label{Tab:PS}
\end{table}

\begin{table*}[ht]
\centering 
\caption{Results ~($\%$) of the state-of-the-art methods on CUB, SUN and AwA2. The best and second-best results are marked in \textbf{bold} and \underline{underline}, respectively. Results are reported in terms of top-1 accuracy of unseen ($U$) and seen ($S$) classes, together with their harmonic mean ($H$).
}\label{Table:SOTA}\vspace{-2mm}
\resizebox{0.9\linewidth}{!}{
\begin{tabular}{r|ccc|ccc|ccc}
\hline
\multirow{2}{*}{\textbf{Methods}\qquad\qquad} 
&\multicolumn{3}{c|}{\textbf{CUB}}&\multicolumn{3}{c|}{\textbf{SUN}}&\multicolumn{3}{c}{\textbf{AwA2}}\\
\cline{2-10} &$U$ &$S$ &$H$ &$U$ &$S$ &$H$ &$U$ &$S$ &$H$ \\ \hline 
\multicolumn{10}{c}{\rule{0pt}{3ex} \textbf{Generative-based Methods}}\\ \hline 
f-VAEGAN (CVPR'19)~\cite{Xian2019FVAEGAND2AF} &48.4&60.1&53.6&45.1&38.0&41.3&57.6&70.6&63.5\\
OCD-CVAE (CVPR'20)~\cite{OCD2020} &44.8&59.9&51.3&44.8&42.9&43.8&59.5&73.4&65.7\\
Composer (NeurIPS'20)~\cite{Huynh2020CompositionalZL} &56.4&63.8&59.9&55.1 &22.0&31.4&62.1&77.3&68.8\\
TF-VAEGAN (ECCV'20)~\cite{TF-VAEGAN2020} &52.8&64.7&58.1&45.6&40.7&43.0&59.8&75.1&66.6 \\
IZF (ECCV'20)~\cite{Shen2020InvertibleZR}&52.7 &68.0 &59.4&52.7 &\underline{57.0} &\underline{54.8}&60.6 &77.5 &68.0 \\
GCM-CF (CVPR'21)~\cite{Yue2021CounterfactualZA}&61.0&59.7&60.3&47.9&37.8&42.2&60.4&75.1&67.0\\
SDGZSL (ICCV'21)~\cite{SDGZSL2021}&59.9&66.4&63.0&--&--&--&64.6&73.6&68.8\\
CE-GZSL (CVPR'21)~\cite{CEGZSL2021}&63.9&66.8&65.3&48.8&38.6&43.1&63.1&78.6&70.0\\
ICCE (CVPR'22)~\cite{ICCE2022} &67.3 &65.5 &66.4 &-- &-- &--&65.3 &82.3 &72.8 \\ 
FREE (ICCV'21)~\cite{FREE2021} &55.7&59.9&57.7&47.4&37.2&41.7&60.4&75.4&67.1\\
HSVA (NeurIPS'21)~\cite{HSVA2021}&52.7&58.3&55.3&48.6&39.0&43.3&59.3&76.6&66.8\\
LBP (TPAMI'21)~\cite{Lu2021ZeroAF}&42.7&71.6 &53.5&39.2 &36.9 &38.1&--&--&--\\
FREE+ESZSL (ICLR'22)~\cite{Cetin2022CL}&51.6&60.4&55.7&48.2&36.5&41.5&51.3&78.0&61.8\\
APN+f-VAEGAN-D2 (IJCV'22)~\cite{Xu2022AttributePN}&65.5&75.6 &70.2&41.4 &\textbf{89.9} &\textbf{56.7}&63.2 &81.0&71.0\\
f-VAEGAN+DSP (ICML'23)~\cite{chen2023evolving}&62.5&73.1 &67.4&57.7 &41.3&48.1&63.7 &88.8 &74.2\\ \hline
\multicolumn{10}{c}{\rule{0pt}{3ex} \textbf{Embedding-based Methods}}\\ \hline 
LDF (CVPR'18)~\cite{LDF2018}&26.4&\textbf{81.6}&39.9&--&--&-- &-- &-- &-- \\
SGMA (NeurIPS'19)~\cite{SGMA2019}&36.7&71.3&48.5&--&--&--&-- &--&--\\
AREN (CVPR'19)~\cite{AREN2019}&63.2&69.0&66.0&40.3&32.3&35.9&54.7&79.1 &64.7 \\
LFGAA (ICCV'19)~\cite{LFGAA2019}&36.2&\underline{80.9}&50.0&18.5&40.0&25.3&27.0&\textbf{93.4}&41.9\\
APN (NeurIPS'20)~\cite{APN2020}&65.3&69.3&67.2&41.9&34.0&37.6&57.1&72.4&63.9\\
DAZLE (CVPR'20)~\cite{DAZLE2020}&56.7&59.6&58.1&52.3&24.3&33.2&60.3&75.7&67.1\\
DVBE (CVPR'20)~\cite{DVBE2020}&53.2&60.2&56.5&45.0&37.2&40.7&63.6&70.8&67.0\\ 
GEM-ZSL (CVPR'21)~\cite{GEM2021}&64.8&77.1&70.4&38.1&35.7&36.9&64.8&77.5&70.6\\
DPPN (NeurIPS'21)~\cite{DPPN2021}&\underline{70.2} &77.1 &73.5 &47.9 &35.8 &41.0&63.1 &86.8 &73.1\\
GNDAN (TNNLS'22)~\cite{Chen2022GNDANGN}&69.2&69.6&69.4&50.0&34.7&41.0&60.2&80.8&69.0\\
MSDN (CVPR'22)~\cite{MSDN2022}&68.7&67.5&68.1&52.2&34.2&41.3&62.0&74.5&67.7\\
TransZero (AAAI'22)~\cite{Chen2021TransZero}&69.3&68.3&68.8&52.6&33.4&40.8&61.3&82.3&70.2\\
TransZero++ (TPAMI'22)~\cite{TransZero++}&67.5&73.6&70.4&48.6&37.8&42.5&64.6&82.7&72.5\\
I2DFormer (NeurIPS'22)~\cite{naeem2022i2dformer}&35.3 &57.6 &43.8&--&--&-- &66.8&76.8&71.5\\
ViT-ZSL (IMVIP'21)~\cite{ViT-ZSL2021}&67.3 &75.2 &71.0&44.5 &55.3 &49.3&51.9 &\underline{90.0} &68.5 \\
IEAM-ZSL (GCPR'21)~\cite{IEAM-ZSL2021}&68.6 &73.8 &71.1 &48.2 &54.7 &51.3&53.7 &89.9 &67.2 \\
DUET (AAAI'23)~\cite{chen2022duet}&62.9 &72.8 &67.5 &45.7 &45.8 &45.8&63.7 &84.7 &72.7 \\
PSVMA (CVPR'2023) (Ours)~\cite{PSVMA2023}&70.1 &77.8 &\underline{73.8}&\textbf{61.7} &45.3 &52.3&\underline{73.6} &77.3 &\underline{75.4}\\
\hline
PSVMA+ (Ours) &\textbf{71.8} &77.8 &\textbf{74.6}&\underline{61.5}&49.4&\underline{54.8} &\textbf{74.2} &86.4 &\textbf{79.8}\\ 
\hline
\end{tabular} }
\label{table:sota}
\end{table*}

\begin{table}[ht]
\tiny
\centering 
\caption{Results ~($\%$) of conventional ZSL. The best and second-best results are marked in \textbf{bold} and \underline{underline}, respectively. 
}\label{Table:SOTAZSL}\vspace{-2mm}
\resizebox{0.9\linewidth}{!}{
\begin{tabular}{r|c|c|c}
\hline
\multirow{2}{*}{Methods\qquad}
&CUB&SUN&AwA2\\
\cline{2-4} 
$ $ &$acc$ & $acc$ & $acc$ \\
\hline  
{GEM-ZSL}~\cite{GEM2021}&\underline{77.8}&62.8&67.3\\
DPPN~\cite{DPPN2021} &\underline{77.8} &61.5&73.3\\
MSDN~\cite{MSDN2022}&76.1&65.8&70.1\\
TransZero~\cite{Chen2021TransZero} &76.8&65.6&70.1\\
TransZero++~\cite{TransZero++} &78.3&67.6&72.6\\
I2DFormer~\cite{naeem2022i2dformer} &45.4 &--&\underline{76.4}\\
DUET~\cite{chen2022duet} &72.3 &64.4 &69.9\\
PSVMA (Ours)~\cite{PSVMA2023} &77.3 &\underline{72.6} &75.2\\
\hline
PSVMA+ (Ours) &\textbf{78.8} &\textbf{74.5} &\textbf{79.2}\\
\hline 
\end{tabular} }
\label{table:sotazsl}
\end{table}

\subsection{Comparison with State-of-the-Art}\label{sec4.1}	
In Table \ref{table:sota}, we provide a comparative analysis of PSVMA+ against recent state-of-the-art methods to demonstrate its effectiveness and advantages. The compared methods are classified into generative-based methods (\textit{e.g.}, TF-VAEGAN~\cite{TF-VAEGAN2020}, IZF~\cite{Shen2020InvertibleZR}, SDGZSL~\cite{SDGZSL2021}, GCM-CF~\cite{Yue2021CounterfactualZA}, FREE~\cite{FREE2021}, HSVA~\cite{HSVA2021}, FREE+ESZSL~\cite{Cetin2022CL}, APN+f-VAEGAN-D2~\cite{Xu2022AttributePN}, f-VAEGAN+DSP~\cite{chen2023evolving}) and embedding-based methods (\textit{e.g.}, APN~\cite{APN2020}, GEM-ZSL~\cite{GEM2021}, DVBE~\cite{DVBE2020}, DPPN \cite{DPPN2021}, GNDAN~\cite{Chen2022GNDANGN}, MSDN ~\cite{MSDN2022}, TransZero~\cite{Chen2021TransZero}, TransZero++ \cite{TransZero++}, ViT-ZSL\cite{ViT-ZSL2021}, IEAM-ZSL \cite{IEAM-ZSL2021}, DUET \cite{chen2022duet}). 

\begin{table*}[t]
\centering
\caption{Conventional zero-shot learning results in 2-hops setting on ImageNet. Results indicated with “*” are taken from TransZero++ \cite{TransZero++}. “$\dagger$” denotes the methods under the same training process.
} \label{table:2hops}
\vspace{-2mm}
\resizebox{1.0\linewidth}{!}{
\begin{tabular}{c|ccccccc}
\hline
Methods &ConSE \cite{Norouzi2014ZeroShotLB}* &SYNC \cite{Changpinyo2016SynthesizedCF}* &EXEM \cite{Changpinyo2017PredictingVE}* &Trivial \cite{Hascoet2019OnZR}* & TransZero++ \cite{TransZero++}*  & TransZero++$\dagger$ & PSVMA+$\dagger$ \\ \hline
Top-1 & 8.3 &10.5 &12.5 &20.3 & 23.9 &17.4 &23.0 \\ \hline 
\end{tabular}}
\end{table*}
\begin{table*}[t]
\centering
\caption{ Ablation studies for different components of PSVMA+ on CUB, SUN, and AwA2 datasets. } \label{table:ablation_component}
\vspace{-2mm}
\resizebox{1.0\linewidth}{!}{
\begin{tabular}{c|cc|cc|c|ccc|c|ccc|c|ccc}
\hline
\multirow{2}{*}{baseline} &\multicolumn{2}{c|}{DSVTM} &\multicolumn{2}{c|}{Multi-granularity} &\multicolumn{4}{c|}{CUB} &\multicolumn{4}{c|}{SUN} &\multicolumn{4}{c}{AwA2} \\ \cline{2-17} 
&IMSE &SMID &SCGL &AMGF &$acc$&$U$ &$S$ &$H$ &$acc$&$U$ &$S$ &$H$ &$acc$&$U$ &$S$ &$H$ \\ \hline
\checkmark &&&&&68.0 &59.8 &68.4 &63.8 &60.2 &43.8 &30.6 &36.0 &67.4 &58.0 &81.6 &67.8 \\
\checkmark &&\checkmark &&&72.9 &67.4 &72.2 &69.7 &67.3 &57.1 &41.0 &47.7 &70.3 &64.8 &74.9 &69.5 \\
\checkmark &\checkmark &\checkmark &&&75.8 &68.8 &76.3&72.4 &72.1 &60.3 &45.3 &51.7&74.0 &72.8 &76.5 &74.6 \\
\checkmark &\checkmark &\checkmark &\checkmark &&77.3 &70.0 &77.0 &73.3 &73.1 &61.0 &48.8 &54.2 &74.5 &71.9 &80.1 &76.2 \\
\checkmark &\checkmark &\checkmark &&\checkmark &76.1 &70.2 &76.4 &73.2 &73.5 &60.1 &47.6 &53.1 &77.4 &70.8 &84.2 &76.9 \\ \hline \hline
\checkmark &\checkmark &\checkmark &\checkmark &\checkmark &\textbf{78.8} &71.8&77.8 &\textbf{74.6}&\textbf{74.5} &61.5&49.4&\textbf{54.8}&\textbf{79.2} &74.2&86.4 &\textbf{79.8} \\
\hline
\end{tabular}}
\end{table*}

Compared to the state-of-the-art methods, PSVMA+ stands out with impressive results, attaining the best harmonic mean $H$ of 74.6\%/79.8\% and second-best $H$ of 54.8\% on CUB/AwA2 and SUN, respectively. Note that the SUN dataset, with only about 16 images per category, presents a unique challenge due to its limited data. As presented in the upper portion of Table \ref{table:sota}, the generative-based methods often exhibit superior performance by data augmentation through the synthesis of visual features of unseen categories. However, in the embedding-based methods (the lower portion of the table), our method still delivers the best results on SUN dataset. Moreover, compared to the methods (\emph{e.g.,} APN\cite{APN2020}, GEM-ZSL\cite{GEM2021}, DPPN\cite{DPPN2021}, MSDN\cite{MSDN2022}, TransZero\cite{Chen2021TransZero}, TransZero++ \cite{TransZero++}) which utilize the sharing attribute prototypes, PSVMA \cite{PSVMA2023} (our conference version) achieves substantial improvements in harmonic mean $H$, with gains exceeding 0.7\%, 9.8\%, and 2.3\% on CUB, SUN, and AwA2, respectively. This demonstrates that the multiple unambiguous representations acquired through instance-centric attributes exhibit greater transferability and discriminative power compared to features guided by sharing attributes. Further, leveraging the hierarchical multi-granularity architecture, PSVMA+ further surpasses its conference version (PSVMA), achieving gains in $H$ of 0.8\%, 2.5\%, and 4.4\% on the three datasets. As shown in the bottom 5 rows of the Table \ref{table:sota}, we also compare PSVMA+ with some ViT-based methods \cite{ViT-ZSL2021,IEAM-ZSL2021,chen2022duet,PSVMA2023}. PSVMA+ consistently achieves the highest $H$ across all datasets. When compared to ViT-ZSL \cite{ViT-ZSL2021} and IEAM-ZSL \cite{IEAM-ZSL2021}, which employ a larger ViT architecture (\emph{i.e.}, ViT-Large), PSVMA+ relies solely on the ViT-Base model, yet it maintains a noteworthy performance advantage. Although IEAM-ZSL is carefully designed to improve the recognition of unseen categories by a self-supervised task, it shows lower performance than PSVMA+ with $U$ falling 3.2\%, 13.3\%, and 10.5\% on CUB, SUN, and AwA2 datasets. 
Additionally, we can see that the seen-unseen performance is not always consistent. Enhancing model transferability (increased $U$) may reduce discrimination (decreased $S$). This is because GZSL methods align with category attributes, but attribute labels of various categories are non-orthogonal to each other. Hence, we pursue a trade-off between the seen and unseen domains to improve overall $H$.

To further investigate the superiority of our method, Table \ref{table:sotazsl} shows the results of our method under the conventional ZSL setting. Our PSVMA+ outperforms the state-of-the-art methods by at least 1.0\%, 1.9\%, 2.8\% on CUB, SUN, and AwA2. These results consistently demonstrate that our PSVMA+ is still effective in conventional ZSL. Additionally, we validate PSVMA+ on ImageNet following \cite{TransZero++}. Table \ref{table:2hops} reports the results in the 2-hops setting. Due to the unavailable source code and computational constraints, we retrain both TransZero++ and PSVMA+ under identical conditions\footnote{Experiments conducted using one 3090 GPU, with 50 epochs, batch size of 64, and learning rate of 1e-4. No multiple iterations of trial and error were performed to set optimal parameters.} (indicated by “$\dagger$”). We randomly set the number of attribute groups as ImageNet does not provide this information. PSVMA+$\dagger$ underperforms the reported TransZero++ by 0.9\%. However, when compared under the same experimental settings, PSVMA+$\dagger$ achieves a 2.6\% improvement over reproduced TransZero++$\dagger$.

\begin{figure*}[t]
\centering
\includegraphics[scale=0.55]{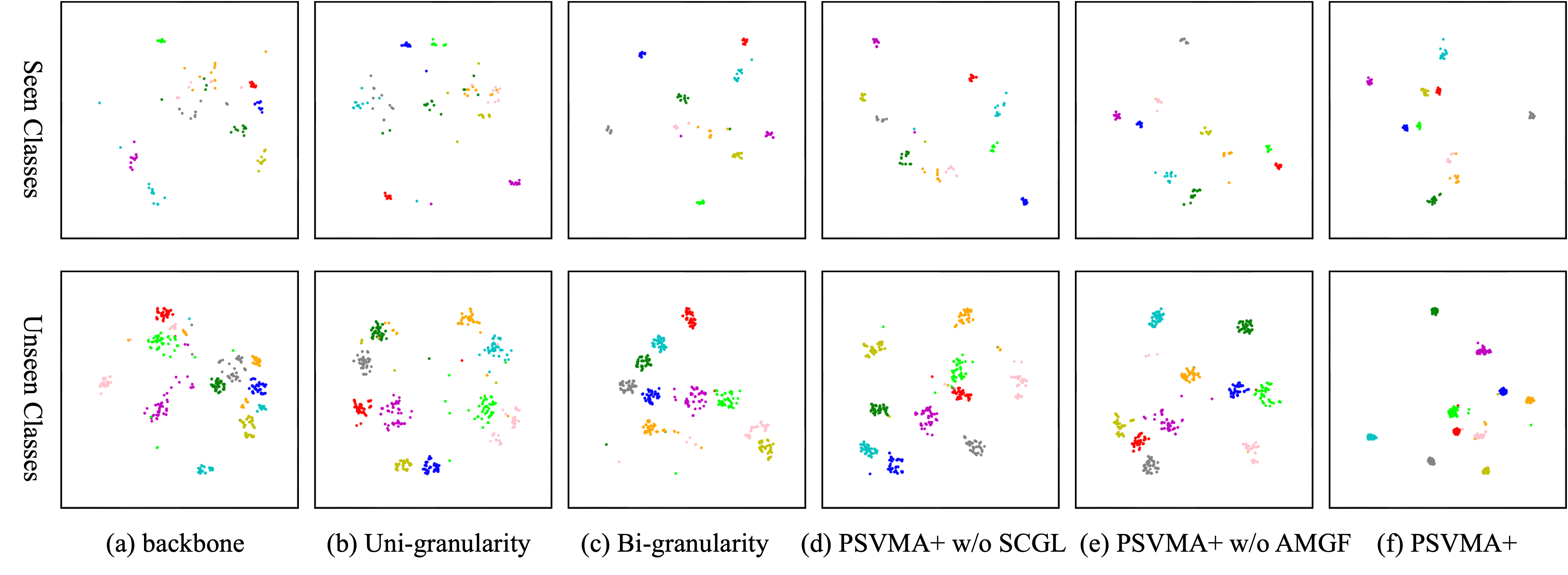}\vspace{-5mm}
\caption{t-SNE visualizations of visual features for seen classes and unseen classes, learned by the (a) ViT backbone, (b) uni-granularity, (c) bi-granularities, (d) PSVMA+ w/o SCGL, (e) PSVMA+ w/o AMGF, and (f) our full PSVMA+. The 10 colors denote 10 different seen/unseen classes randomly selected from CUB.}
\label{fig:tsne}
\end{figure*}

\subsection{Ablation Study}\label{sec4.2}
To give a clear insight into our framework, we perform ablations to analyze the effectiveness of DSVTM and multi-granularity learning.

\begin{figure*}[t]
\begin{center}
\includegraphics[scale=0.5]{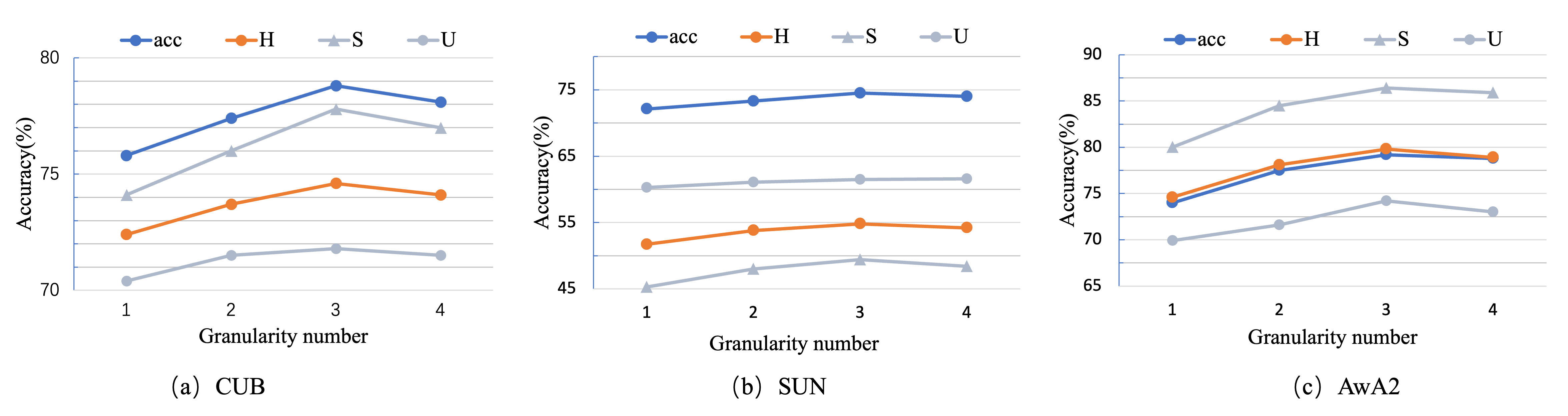}
\vspace{-2mm}
\caption{Effect of granularity number on (a) CUB, (b) SUN, and (c) AwA2 datasets.}
\label{fig:Different Granularity}
\end{center}
\end{figure*} 

\begin{table*}[t]
\centering
\caption{ Ablation studies for different optimizing functions for the granularity-aware loss in PSVMA+ on CUB, SUN, and AwA2 datasets.} \label{table:ablation_ge}
\vspace{-2mm}
\resizebox{1.0\linewidth}{!}{
\begin{tabular}{c|ccccc|c|ccc|c|ccc|c|ccc}
\hline
\multirow{2}{*}{} &\multicolumn{5}{c|}{SCGL} &\multicolumn{4}{c|}{CUB} &\multicolumn{4}{c|}{SUN} &\multicolumn{4}{c}{AwA2} \\ \cline{2-18} 
&$\ell_1$ &$\ell_2$ &KL &JSD &ours &$acc$&$U$ &$S$ &$H$ &$acc$&$U$ &$S$ &$H$ &$acc$&$U$ &$S$ &$H$ \\ \hline
\multirow{5}{*}{\rotatebox {90}{PSVMA+}} &\checkmark &&&&&76.2 &68.4 &76.3 &72.1 &72.6 &61.5 &46.2 &52.8 &77.6 &72.8 &81.8 &77.2\\
&&\checkmark &&&&76.4 &67.9 &77.6 &72.5 &73.4 &62.4 &46.6 &53.3 &78.0 &73.0 &84.6 &78.4 \\
& & &\checkmark &&&77.5 &70.3 &76.3 &73.2 &73.8 &60.4 &48.6 &53.9 &78.9 &73.2 &84.6 &78.5\\
& & & & \checkmark & &76.0 &68.4 &77.4 &72.6 &72.6 &61.3 &46.1 &52.6 &77.7 &72.5 &83.6 &77.7 \\
& & & & &\checkmark &\textbf{78.8} &71.8 &77.8 &\textbf{74.6} &\textbf{74.5} &61.5 &49.4 &\textbf{54.8} &\textbf{79.2} &74.2 &86.4 &\textbf{79.8} \\ \hline 
\end{tabular}}
\end{table*}

\begin{table}[t]
\centering
\caption{Ablation studies for $\ell_1$ and $\ell_2$ loss in Eq (19) on CUB, SUN, and AwA2 datasets.} \label{table:ablation_eq19}
\tabcolsep=0.1cm
\resizebox{3.5in}{!}{
\begin{tabular}{cc|c|ccc|c|ccc|c|ccc}
\hline
\multicolumn{2}{c|}{Eq (19)} &\multicolumn{4}{c|}{CUB} &\multicolumn{4}{c|}{SUN} &\multicolumn{4}{c}{AwA2} \\ \cline{1-14} 
$\ell_1$ &$\ell_2$ &$acc$&$U$ &$S$ &$H$ &$acc$&$U$ &$S$ &$H$ &$acc$&$U$ &$S$ &$H$ \\ \hline
\checkmark & &\textbf{78.9} &70.9 &78.7 &\textbf{74.6} &74.2 &61.1 &49.5 &54.7 &79.0 &73.9 &86.6 &79.7 \\
&\checkmark &78.8 &71.8 &77.8 &\textbf{74.6} &\textbf{74.5} &61.5 &49.4 &\textbf{54.8} &\textbf{79.2} &74.2 &86.4 &\textbf{79.8} \\ \hline 
\end{tabular}}
\end{table}

\subsubsection{Effect of IMSE and SMID in DSVTM}
Table \ref{table:ablation_component} summarizes the results of ablation studies. Firstly, the baseline (1st row) refers to the approach in which we directly compute the similarity scores between the visual features extracted from ViT and the category prototypes to infer the category.
Compared to the baseline, SMID (2nd row) brings $acc$/$H$ gains of 4.9\%/5.9\%, 7.1\%/11.7\%, and 2.9\%/1.7\% on CUB, SUN, and AwA2 datasets, respectively, which demonstrates the effectiveness of SMID for semantic-related instance adaption. After that, we incorporate IMSE into this model (3rd row) to learn instance-centric semantic attributes rather than the sharing attributes. Therefore, the model can get the $acc$/$H$ improvements of 2.9\%/2.7\%, 4.8\%/4.0\%, and 3.7\%/5.1\% on CUB, SUN, and AwA2 datasets, respectively, benefiting from semantic-visual mutual adaptation.

\subsubsection{Investigation of Multi-Granularity Learning}
\noindent\textbf{Effect of SCGL and AMGF.}
To leverage the granularity disparities, we propose the exploration of multi-granularity learning via SCGL and AMGF to significantly enhance the transferability and discriminability of our model. SCGL (4th row) achieves gains of 1.5\%/0.9\%, 1.0\%/2.5\%, and 0.5\%/1.6\% in $acc$/$H$ on CUB, SUN, and AwA2 respectively. This indicates the importance of our SCGL that learns cross-granularity representations for GZSL. Besides, AMGF (5th row) increases the performance of PSVMA+ by adaptively fusing the multi-granularity features to discover distinguishing comprehensive features for making decisions. Combined with the SCGL and AMGF, our PSVMA+ (6th row) achieves a substantial gain margin compared to the baseline, \textit{i.e.,} 20.8\%/10.8\%, 14.3\%/18.8\%, and 11.8\%/12.0\% in $acc$/$H$ on CUB, SUN, and AwA2 datasets, respectively.

\noindent\textbf{t-SNE Visualizations.}
In Fig. \ref{fig:tsne}, we present a t-SNE visual comparison of the features from both seen and unseen categories, obtained from (a) the ViT backbone, (b) the uni-granularity approach, (c) the bi-granularity model, (d) PSVMA+ w/o SCGL, (e) PSVMA+ w/o AMGF, and (f) our complete PSVMA+. We observe that features acquired directly from the ViT backbone are ineffective for GZSL, posing a challenge in the identification of classes from the seen and unseen sets. Comparing uni-granularity (\emph{cf} Fig. \ref{fig:tsne} (b)) to multi-granularity (bi-granularity and tri-granularity in Fig. \ref{fig:tsne} (c) and (f)), the latter exhibits a more compact and discriminative representation, effectively mitigating granularity discrepancies and boosting the category recognition performance. As for the proposed SCGL and AMGF, we can see that removing any one of them leads to a distinguishing performance drop (\emph{cf} Fig. \ref{fig:tsne} (d) and (e)). Fig. \ref{fig:tsne} (f) suggests that combining within a unified framework of SCGL and AMGF shows discriminative capability with higher inter-class discrepancy and clearer decision boundaries. Consequently, our PAVMA+ achieves optimal recognition capabilities for both seen and unseen categories.

\noindent\textbf{Influence of Granularity.}
We study the number of granularities in our hierarchical multi-granularity architecture and give the recognition results in Fig. \ref{fig:Different Granularity}. As can be seen, an increase in the number of granularities results in significant performance improvements compared to uni-granularity. For instance, on the CUB, SUN, and AwA2 datasets, we observe increases of at least 1.6\%/ 1.3\%, 1.2\%/ 2.1\%, and 3.5\%/ 3.5\% in $acc$/$H$, respectively. This demonstrates the effectiveness of multi-granularity learning, as it gathers more visual clues from different granularities, ultimately facilitating a comprehensive representation capable of distinguishing categories across a variety of GZSL datasets. Nevertheless, the introduction of more granularity does not consistently lead to improvements, which can even cause slight accuracy degradation when the number of granularities exceeds 3. This may be attributed to the inclusion of overly shallow visual and semantic features in the hierarchical multi-granularity architecture, introducing information redundancy and confusion, thereby impeding the learning of discriminative knowledge.

\noindent\textbf{Analysis of the Optimization for SCGL}.
Given the diverse contributions of different granularities, SCGL utilizes the granularity-aware KL loss $\hat{D}_{K L}$ to guide the refinement of unreliable granularity and challenging samples. To investigate its effect, we substitute it with several optimization functions, including $\ell_1$, $\ell_2$, KL, JSD. Table \ref{table:ablation_ge} presents the evaluation results on the CUB, SUN, and AwA2 datasets. Compared to $\hat{D}_{K L}$, $\ell_1$ and $\ell_2$ exhibit limited effectiveness in cross-granularity learning, resulting in a decrease in performance, \textit{i.e.,} the $acc$/$H$ drops by at least 2.4\%/3.9\%, 1.1\%/1.5\%, and 1.2\%/1.4\% on CUB, SUN, and AwA2 respectively. In addition, the KL loss proves to be more effective than JSD, indicating that our enhancement of granularity features places a stronger emphasis on asymmetry and a greater reliance on a reliable teacher granularity to guide the student granularity. Moreover, by placing emphasis on difficult samples within the KL loss, we have crafted our loss function $\hat{D}_{K L}$, ultimately enabling PSVMA+ to reach its optimal performance.


\begin{figure*}[t]
\begin{center}
\includegraphics[scale=0.63]{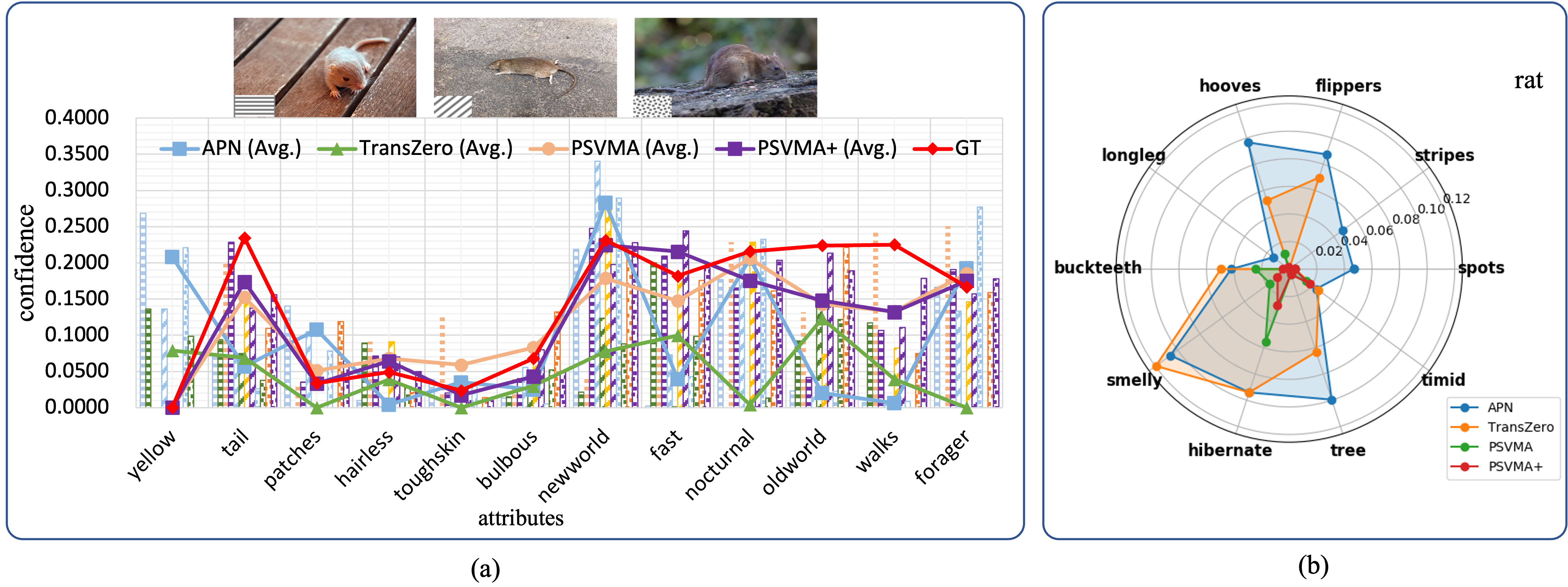}
\caption{Visualization for attribute disambiguation. (a) Semantic disambiguation. The bar with a different mark corresponds to the image with the same mark. The blue, green, orange, and purple bars denote the results of APN, TransZero, PSVMA, and PSVMMA+, respectively. The line graph represents the average results of three randomly selected images. (b) Attribute prediction error that represents the absolute errors between the predicted confidences and the GT of attributes.}
\label{fig:disam}
\end{center}
\end{figure*}

\begin{figure*}[t]
\centering
\includegraphics[scale=0.43]{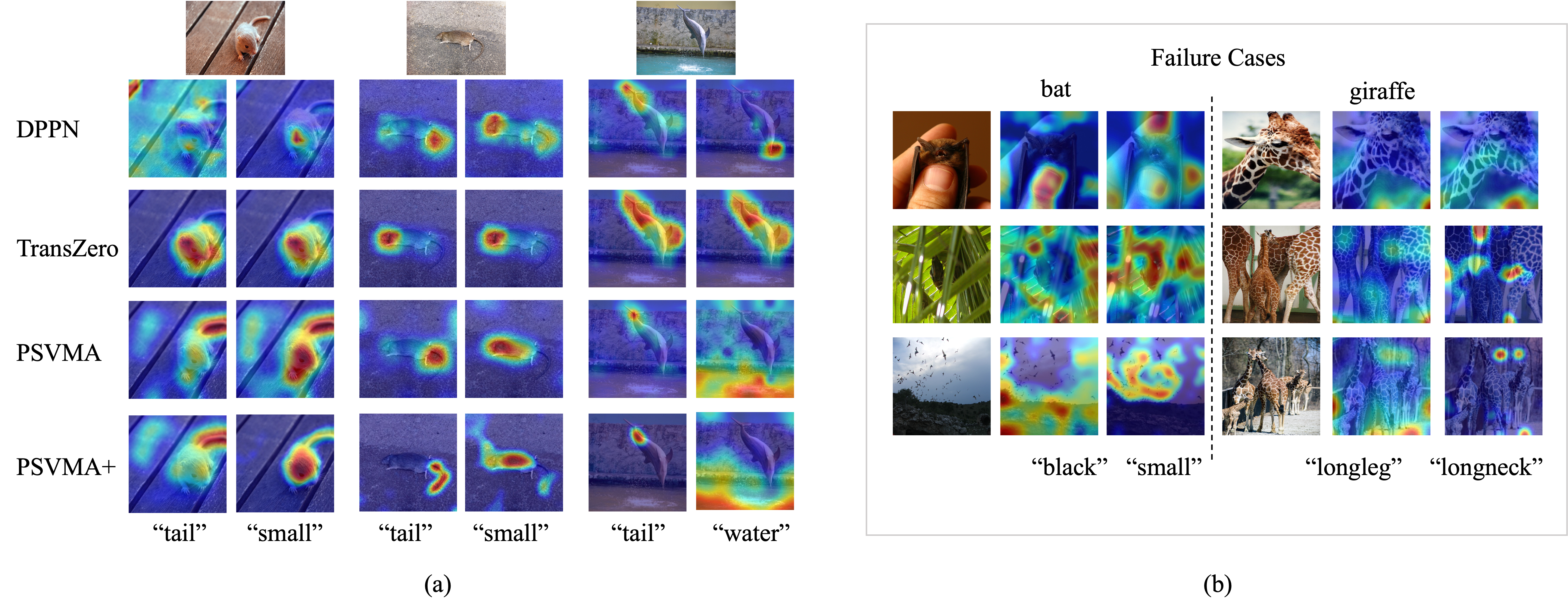}
\caption{Visualization of attention maps. (a) Attention maps of DPPN \cite{DPPN2021}, PSVMA \cite{PSVMA2023} (Conference version), and our PSVMA+. (b) Some failure cases of our PSVMA+.}
\label{fig:atten}
\end{figure*}

\subsubsection{Analysis of Norms in $\mathcal{L}_{deb}$}
To investigate the impact of different norms in $\mathcal{L}_{deb}$ (Eq. (\ref{eqlbias})), we conducted experiments comparing the $\ell_2$ norm with the $\ell_1$ norm. As shown in Table \ref{table:ablation_eq19}, both $\ell_1$ and $\ell_2$ norms yield comparable performance. The differences in $acc$ and $H$ are 0.3 or less for all datasets, indicating the robustness of the model. While the performance differences are minimal, we ultimately retained the $\ell_2$ norm in our final model due to its better overall $H$ performance. However, our analysis indicates that either norm could be used without significantly affecting the results.

\subsection{Qualitative Results}\label{sec4.3}
For GZSL, it is intractable to obtain qualitative results. Here we present the visualization of semantic disambiguation and attention maps, so as to aid an intuitive understanding of PSVMA+’s strength.

\begin{figure*}[t]
\begin{center}
\includegraphics[scale=0.5]{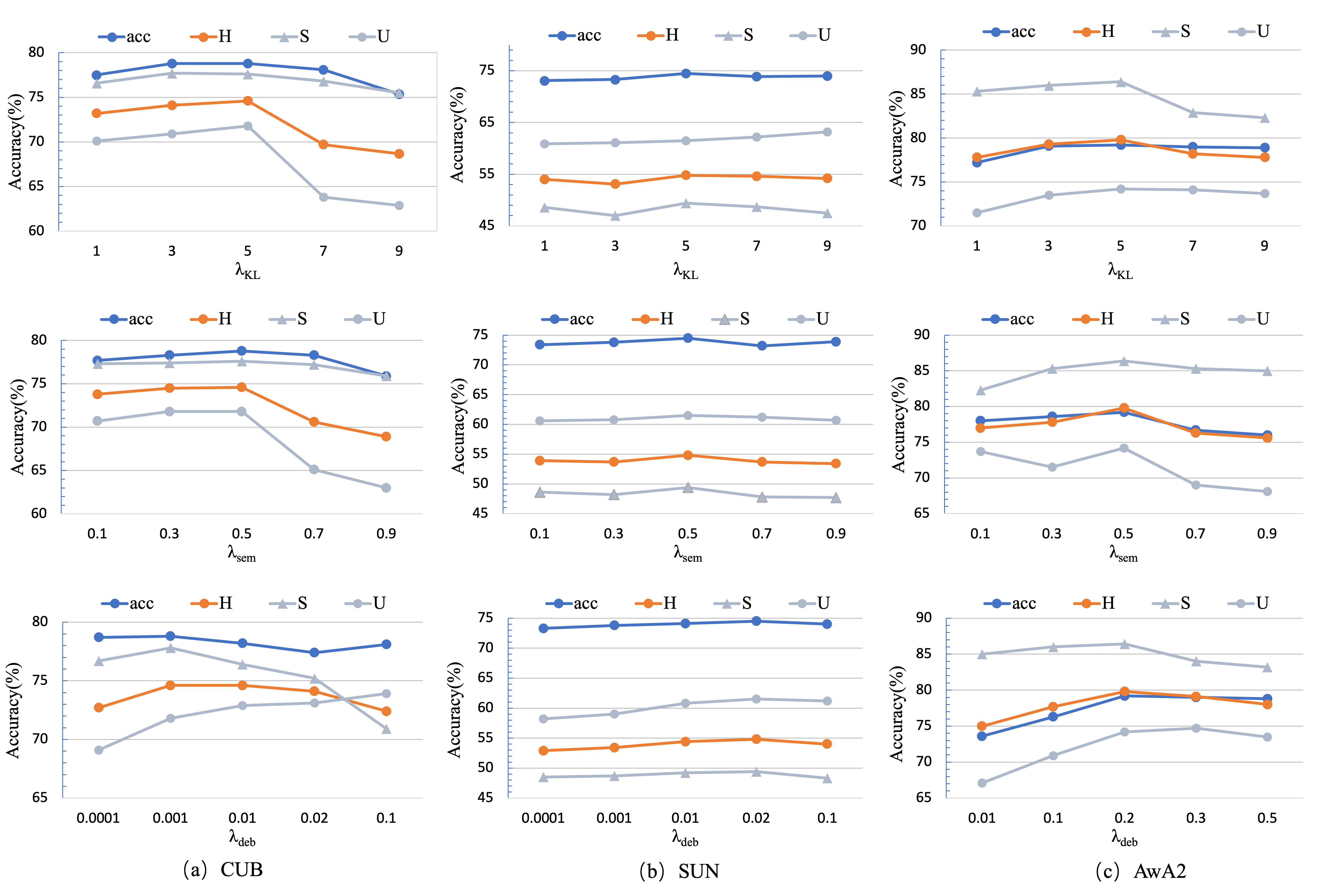}
\caption{Effect of loss weights on (a) CUB (b) SUN and (c) AwA2 datasets.}
\label{fig:lossw}
\end{center}
\end{figure*}

To investigate the semantic disambiguation ability of our method, we calculate the predicted probability of category attribute ($\phi(F_A)$ in Eq. (\ref{eq:score})) as the confidence and compare with several methods including APN\cite{APN2020}, TransZero\cite{Chen2021TransZero}, PSVMA \cite{PSVMA2023}. These methods all use sharing attributes and have the same attribute prediction and category decision formulas. As shown in Fig. \ref{fig:motivation} (e), the attribute ``tail'' shows different appearances in a dolphin's and a rat's image (the red box). APN and TransZero fail to infer the ``tail'' in the dolphin, while our method predicts the attribute in both of the dolphin and rat correctly. Additionally, visual discrepancies for the same attribute information occur not only between classes but also within a class, especially for non-rigid objects with variable postures. Fig. \ref{fig:disam} (a) gives some attribute predictions of three randomly selected dolphin images. In the three intra-class instances, our method successfully recognizes that the rats do not possess the “yellow” property, instead showing strong probabilities of being ``nocturnal'', ``fast'', and ``new world''. Overall, the average attribute predictions of the three images are more consistent with the GT (ground truth) compared to the APN and TransZero methods. For a clearer depiction of the differences between predicted probability and GT, we employ a radial chart (\emph{cf} Fig. \ref{fig:disam} (b)) to visualize prediction errors on more attributes. In comparison to the APN and TransZero, the points of our PSVMA+ are closer to the center of the chart, signifying a closer alignment with GT. When compared to PSVMA \cite{PSVMA2023} (our conference version), PSVMA+ predicts attribute confidences more precisely and possesses superior inter-class and intra-class attribute disambiguation. These demonstrate that the multi-granularity exploration of semantic-visual adaption is beneficial for the collection of sufficient visual clues for semantic knowledge transfer, leading to better attribute prediction ability.

Additionally, we visualize the attention maps learned by PSVMA+ and other methods \cite{Chen2021TransZero,DPPN2021,PSVMA2023}. As shown in Fig. \ref{fig:atten} (a), for the rat and dolphin images, our method successfully identifies relevant objects of the given attribute ``tail'', while other methods provide less precise localization and neglect critical features~\cite{DPPN2021,Chen2021TransZero}.
This demonstrates the effectiveness of matching semantic-visual pairs in alleviating both inter-class and intra-class semantic ambiguities. When DPPN struggles to identify the ``water'' attribute and incorrectly associates it with the head area, our approach captures visual details corresponding to the ``water''. Furthermore, compared to PSVMA, PSVMA+ excels in identifying attribute-related regions with finer details. Taking the second rat image as an example, our approach provides a more specific localization of the tail shape, rather than a mere point. We may give credit to the equipment of granularity-wise semantic-visual interaction. Fig. \ref{fig:atten} (b) also presents some failure cases where our model incorrectly localizes target attributes. These cases illustrate challenges with occlusions, incomplete objects, and multiple objects. As demonstrated in rows 1-2 of the bat images and row 1 of the giraffe images, the model often incorrectly assigns attributes to the occluding object. Furthermore, when an image contains multiple objects, the model sometimes becomes confused. In such cases, it may incorrectly localize attributes to the background or confuse different attributes. For instance, in rows 2-3 of the giraffe images, the model struggles to differentiate between the ``longleg'' and ``longneck'' attributes.

\subsection{Hyperparameter Analysis}\label{sec4.6}	
PSVMA+ utilizes loss weights $\lambda _{sem}$, $\lambda_{KL}$, and $\lambda_{deb}$ to control their corresponding loss terms (\textit{i.e.}, $\mathcal{L}_{sem}$, $\hat{D}_{K L}$, $\mathcal{L}_{deb}$) from Eq. (\ref{eq:loss}). To obtain the optimal parameters, we evaluate the effects of these weights on CUB, SUN, and AwA2 datasets (see Fig. \ref{fig:lossw}). As $\lambda _{sem}$ increases, both $acc$ and $H$ rise slowly at first and then decrease when $\lambda _{sem} >0.5$ on all datasets. A large value of $\lambda _{sem}$ over-emphasizes the knowledge on the seen domain by Eq. (\ref{eq:L_sem}), resulting in inferior generalization capability. As a result, we opt for $\lambda _{sem}=0.5$ for the CUB, SUN, and AwA2 datasets. For $\lambda_{KL}$, we explore a large range due to the high temperature introduced during the knowledge distillation process. Then, we gradually increase the value of $\lambda _{deb}$. For the CUB dataset, we chose $\lambda _{deb} = 0.001$, which differs from the values used in the other two datasets. This choice is driven by the unique characteristics of the CUB dataset, which contains fine-grained categories of bird species with high intra-class similarity. Thus, it requires a more nuanced regularization to effectively distinguish between subtly different classes. When more attention is paid to pursuing the distribution consistency between seen and unseen predictions, we get better unseen performance. Nevertheless, all $\lambda _{sem}$, $\lambda _{KL}$, and $\lambda _{deb}$ can not be too large to avoid squeezing the capacity of classification loss, resulting in recognition accuracy reduction. Based on these observations, we fix $\{\lambda _{sem}$, $\lambda_{KL}$, $\lambda_{deb}\}$ to $\{0.5, 5.0, 0.001\}$, $\{0.5, 5.0, 0.02\}$, $\{0.5, 5.0, 0.2\}$ for CUB, SUN, and AwA2, respectively. 

\subsection{Computational Analysis}
\begin{table}[t]
\centering
\caption{Comparison of computational complexity between DPPN, PSVMA and PSVMA+ on AwA2 datasets.}
\label{tab:Comparisoncomparison}
\begin{tabular}{l|c|c|c}
\hline
\textbf{Methods} &DPPN& \textbf{PSVMA} & \textbf{PSVMA+}\\
\hline
Performance ($Acc$) &73.3 & 75.2 & \textbf{79.2} \\
Performance ($H$) &73.1 & 75.4 & \textbf{79.8} \\
GFLOPs &32.0 & \textbf{21.8} & 31.5\\
\hline
\end{tabular}
\end{table}
In Table \ref{tab:Comparisoncomparison}, we compare computational complexity with our previous work (PSVMA\cite{PSVMA2023}) and DPPN \cite{DPPN2021} (the previous best method on AwA2) by reporting the number of GFLOPs. As shown in the table, PSVMA exhibits the lowest GFLOPs and better ZSL/GZSL performance than DPPN. PSVMA+ involves higher computational cost as it adopts a hierarchical architecture containing multiple DSVTM structures. This increase in complexity brings significant performance improvements in both ZSL (4.0 $\uparrow$) and GZSL (4.4 $\uparrow$). Nevertheless, PSVMA+ is still superior to DPPN in terms of both effectiveness (ZSL: 79.2 v.s. 73.3) and efficiency (31.5 v.s. 32.0 GFLOPs).

\section{Conclusion}\label{sec5}
This work delves into multi-granularity progressively semantic-visual adaption (PSVMA+) and uncovers the insights of insufficient correspondence caused by the diversity of attributes and instances. To capture consistent semantic-visual alignment, we propose dual semantic-visual transformer modules (DSVTM) which granularity-wisely reveal and gather adequate visual clues for various attributes, exploiting the granularity disparities and facilitating semantic knowledge transfer. DSVTM adapts the sharing attributes into instance-centric attributes and aggregates semantic-related visual regions, which learn unambiguous visual representation to accommodate various visual instances. 
With hierarchical DSVTMs, the visual objects are injected into different levels to match and pinpoint different attributes, yielding granularity-specific features. Due to the diverse contributions of different granularities, PSVMA+ utilizes selective cross-granularity learning (SCGL) to transfer the discriminative knowledge from reliable granularity to improve the representations of the unreliable one and challenging samples. Moreover, adaptive multi-granularity fusion (AMGF) combines features to collaborate with each granularity and obtain a comprehensive and distinguishable representation, improving the transferability and discriminability. Extensive experiments on three public datasets show the superiority of our PSVMA+.

\section{ACKNOWLEDGMENTS}
This work was supported in part by the National Key R\&D Program of China (No.2021ZD0112100), National Natural Science Foundation of China (No. 62331003, 62120106009, 62302141, 62072026, 62476021), Beijing Natural Science Foundation (L223022), Natural Science Foundation of Hebei Province (F2024105029), Fundamental Research Funds for the Central Universities (JZ2024HGTB0255), and the Chinese Association for Artificial Intelligence (CAAI)-Compute Architecture for Neural Networks (CANN) Open Fund, developed on OpenI Community.

\ifCLASSOPTIONcaptionsoff
\newpage
\fi

\bibliographystyle{IEEEtran}
\bibliography{mybibfile}

\end{CJK}

\begin{IEEEbiography}
	[{\includegraphics[width=1in,height=1.25in,clip,keepaspectratio]{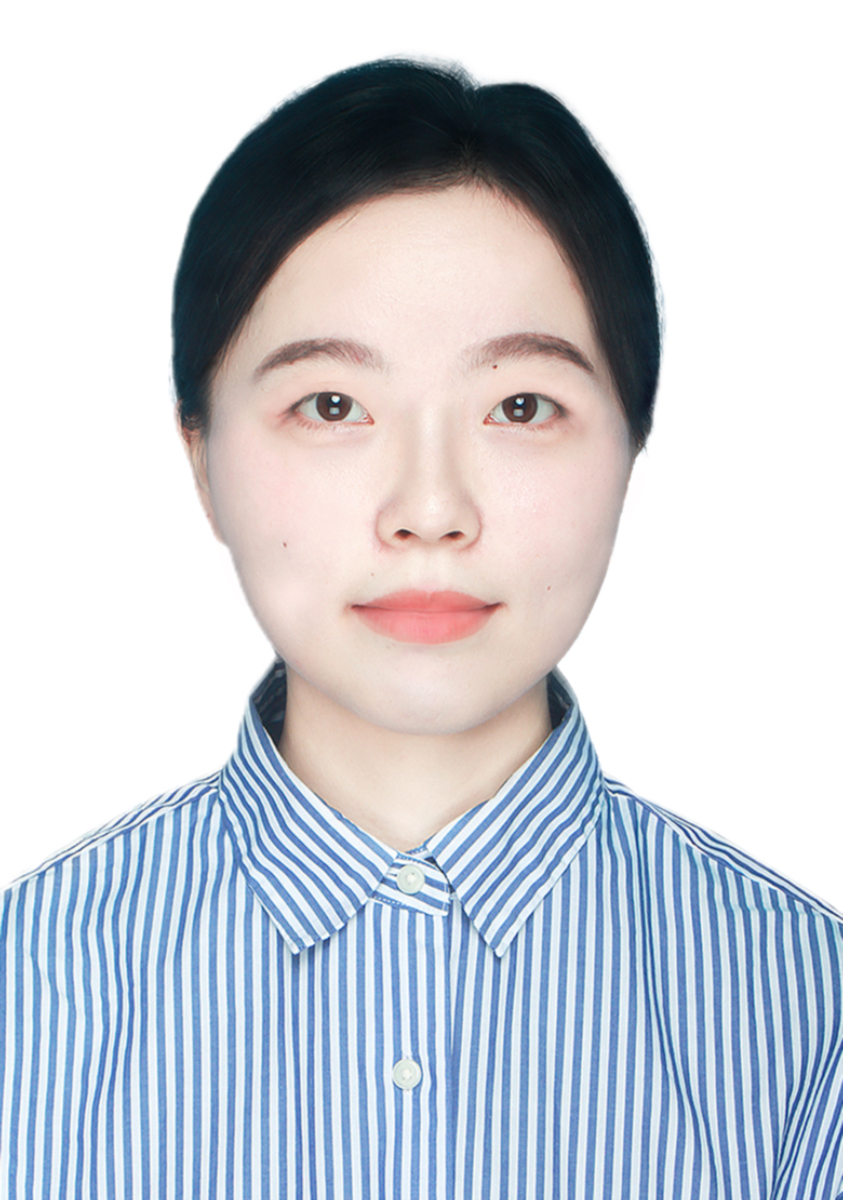}}]
	{Man Liu}
	received the M.S. degree in pattern recognition and intelligent system from the Anhui University of Technology, Ma'anshan, China, in 2020. She is currently pursuing the Ph.D. degree in control science and engineering with the Institute of Information Science, Beijing Jiaotong University, Beijing, China. Her current research interests include image classification, zero-shot learning, visual-and-language learning.
\end{IEEEbiography}

\begin{IEEEbiography}
	[{\includegraphics[width=1in,height=1.25in,clip,keepaspectratio]{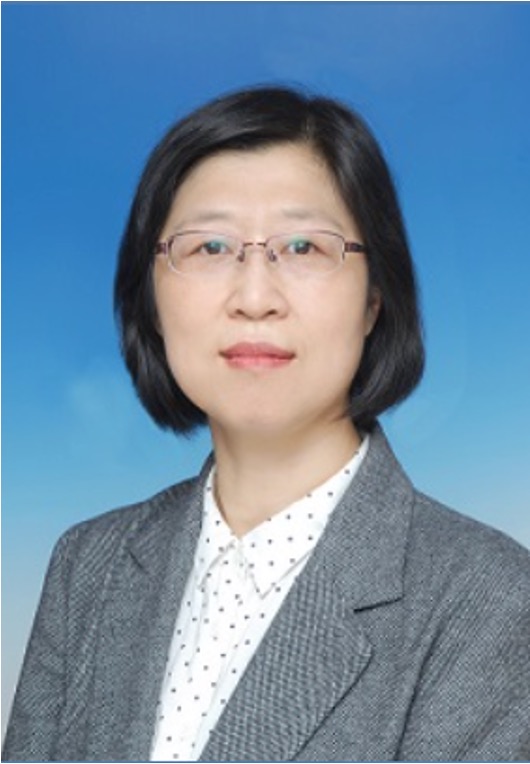}}]
	{Huihui Bai}
	received her B.S. degree from Beijing Jiaotong University, China, in 2001, and her Ph.D. degree from Beijing Jiaotong University, China, in 2008. She is currently a professor in Beijing Jiaotong University. She has been engaged in R \& D work in video coding technologies and standards, such as HEVC, 3D video compression, multiple description video coding (MDC), and distributed video coding (DVC).
\end{IEEEbiography}

\begin{IEEEbiography}
	[{\includegraphics[width=1in,height=1.25in,clip,keepaspectratio]{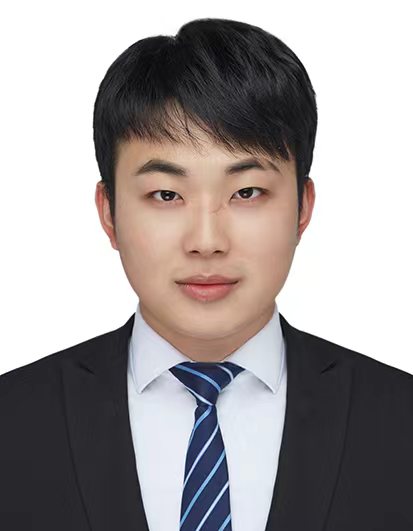}}]
	{Feng Li}
	received his B.S. degree in Anhui Normal University, China, in 2016, and his Ph.D. degree from Beijing Jiaotong University, China, in 2022. He is currently an Associate Professor with the School of Computer Science and Information Engineering, Hefei University of Technology, Hefei, China. His research interests are in image and video compression, image and video super-resolution, and other low-level computer vision tasks, with the focus on deep-learning-based methods.
\end{IEEEbiography}

\begin{IEEEbiography}
	[{\includegraphics[width=1in,height=1.25in,clip,keepaspectratio]{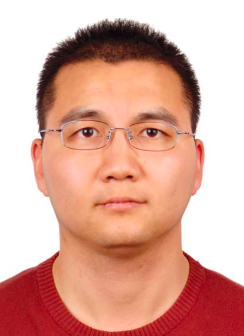}}]
	{Chunjie Zhang}
	received the Ph.D. degree from the National Laboratory of Pattern Recognition, Institute of Automation, Chinese Academy of Sciences, China, in 2011. He worked as an Engineer at the Henan Electric Power Research Institute from 2011 to 2012. He worked as a Postdoctoral Researcher at the School of Computer and Control Engineering, University of Chinese Academy of Sciences, Beijing, China. Then, he joined the School of Computer and Control Engineering, University of Chinese Academy of Sciences, as an Assistant Professor. In 2017, he joined the Institute of Automation, Chinese Academy of Sciences, as an Assistant Professor, where he was then promoted to an Associate Professor in October 2017. In 2019, he joined the Institute of Information Science, Beijing Jiaotong University, as a Professor. His current research interests include image processing, cross-media analysis, machine learning, pattern recognition, computer vision, and applications in rail traffic control and safety.
\end{IEEEbiography}

\begin{IEEEbiography}[{\includegraphics[width=1in,height=1.25in,clip,keepaspectratio]{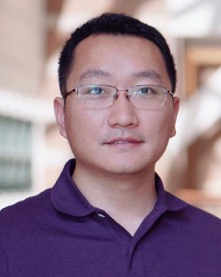}}]{Yunchao Wei}
is currently a professor at the Center of Digital Media Information Processing, Institute of Information Science, at Beijing Jiaotong University. He received his Ph.D. degree from Beijing Jiaotong University, Beijing, China, in 2016. He was a Postdoctoral Researcher at Beckman Institute, UIUC, from 2017 to 2019.
He is ARC Discovery Early Career Researcher Award Fellow from 2019 to 2021. His current research interests include computer vision and
machine learning.
\end{IEEEbiography}

\begin{IEEEbiography}[{\includegraphics[width=1in,height=1.25in,clip,keepaspectratio]{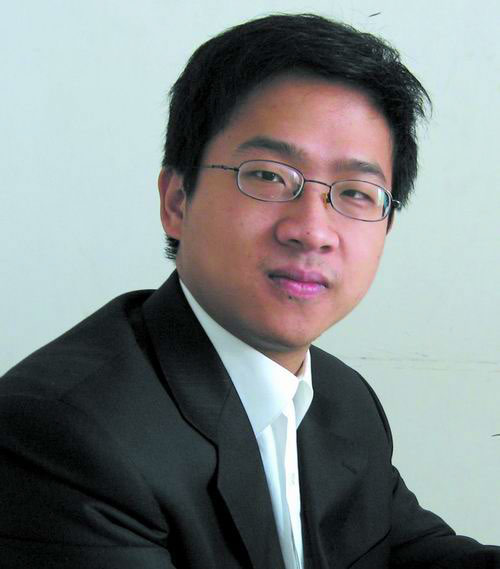}}]{Meng Wang}
(Fellow, IEEE) received the B.E. and Ph.D. degrees from the Special Class for the Gifted Young, Department of Electronic Engineering and Information Science, University of Science and Technology of China (USTC), Hefei, China, in 2003 and 2008, respectively. He is currently a Professor with the Hefei University of Technology, China. He received paper prizes or awards from ACM MM 2009 (the Best Paper Award), ACM MM 2010 (the Best Paper Award), ACM MM 2012 (the Best Demo Award), ICDM 2014 (the Best Student Paper Award), SIGIR 2015 (the Best Paper Honorable Mention), IEEE TMM 2015 and 2016 (the Prize Paper Award Honorable Mention), IEEE SMC 2017 (the Best Transactions Paper Award), and ACM TOMM 2018 (the Nicolas D. Georganas Best Paper Award). He is currently an Associate Editor of IEEE Transactions on Pattern Analysis and Machine Intelligence, IEEE Transactions on Knowledge and Data Engineering, IEEE Transactions on Multimedia, and IEEE Transactions on Neural Networks and Learning Systems.
\end{IEEEbiography}

\begin{IEEEbiography}[{\includegraphics[width=1in,height=1.25in,clip,keepaspectratio]{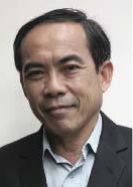}}]{Tat-Seng Chua} is the KITHCT Chair Professor at the School of Computing, National University of Singapore (NUS). He is also the Distinguished Visiting Professor of Tsinghua University, the Visiting Pao Yue-Kong Chair Professor of Zhejiang University, and the Distinguished Visiting Professor of Sichuan University. Dr. Chua was the Founding Dean of the School of Computing from 1998-2000. His main research interests include unstructured data analytics, video analytics, conversational search and recommendation, and robust and trustable AI. He is the Co-Director of NExT, a joint research Center between NUS and Tsinghua University, and Sea-NExT, a joint Lab between Sea Group and NExT.
Dr. Chua is the recipient of the 2015 ACM SIGMM Achievements Award, and the winner of the 2022 NUS Research Recognition Award. He is the Chair of steering committee of Multimedia Modeling (MMM) conference series, and ACM International Conference on Multimedia Retrieval (ICMR) (2015-2018). He is the General Co-Chair of ACM Multimedia 2005, ACM SIGIR 2008, ACM Web Science 2015, ACM MM-Asia 2020, and the upcoming ACM conferences on WSDM 2023 and TheWebConf 2024. He serves in the editorial boards of three international journals. Dr. Chua is the co-Founder of two technology startup companies in Singapore. He holds a PhD from the University of Leeds, UK.
\end{IEEEbiography}

\begin{IEEEbiography}
	[{\includegraphics[width=1in,height=1.25in,clip,keepaspectratio]{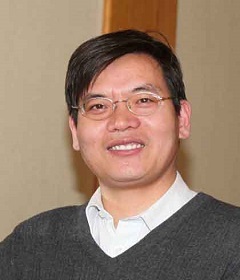}}]
	{Yao Zhao}
	(Fellow, IEEE) received the B.S. degree from the Radio Engineering Department, Fuzhou University, Fuzhou, China, in 1989, the M.E. degree from the Radio Engineering Department, Southeast University, Nanjing, China, in 1992, and the Ph.D. degree from the Institute of Information Science, Beijing Jiaotong University (BJTU), Beijing, China, in 1996, where he became an Associate Professor and a Professor in 1998 and 2001, respectively. From 2001 to 2002, he was a Senior Research Fellow with the Information and Communication Theory Group, Faculty of Information Technology and Systems, Delft University of Technology, Delft, The Netherlands. In 2015, he visited the Swiss Federal Institute of Technology, Lausanne (EPFL), Switzerland. From 2017 to 2018, he visited University of Southern California. He is currently the Director with the Institute of Information Science, BJTU. His current research interests include image/video coding, digital watermarking and forensics, video analysis and understanding, and artificial intelligence. Dr. Zhao is a Fellow of the IET. He serves on the Editorial Boards of several international journals, including as an Associate Editor for the IEEE Transactions on Cybernetics, IEEE Signal Processing Letters, and an area editor of Signal Processing: Image Communication (Elsevier), etc. He was named a Distinguished Young Scholar by the National Science Foundation of China in 2010 and was elected as a Chang Jiang Scholar of Ministry of Education of China in 2013. 
\end{IEEEbiography}
\end{document}